\documentclass[lettersize,journal]{IEEEtran}
\usepackage{amsmath,amsfonts}
\usepackage{algorithmic}
\usepackage{algorithm}
\usepackage{array}
\usepackage[caption=false,font=normalsize,labelfont=sf,textfont=sf]{subfig}
\usepackage{textcomp}
\usepackage{stfloats}
\usepackage{url}
\usepackage{verbatim}
\usepackage{graphicx}
\usepackage{cite}
\usepackage{gensymb}
\usepackage{hyperref}
\usepackage{booktabs}       
\usepackage{multirow}
\usepackage{bm}

\usepackage{xcolor}

\begin{document}

\title{FedRSU: Federated Learning for Scene Flow Estimation on Roadside Units}

\author{
Shaoheng Fang$^{\dagger}$, Rui Ye$^{\dagger}$, Wenhao Wang$^{\dagger}$, Zuhong Liu, Yuxiao Wang, Yafei Wang~\IEEEmembership{Member},  Siheng Chen~\IEEEmembership{Senior Member}, Yanfeng Wang

\thanks{$\dagger$: equal contribution.}
\thanks{Shaoheng Fang, Rui Ye, Yuxiao Wang are with the Cooperative Medianet Innovation Center at Shanghai Jiao Tong University, Shanghai, China (Email: shfang@sjtu.edu.cn, yr991129@sjtu.edu.cn, yuxiao9513@gmail.com ).}
\thanks{Wenhao Wang is with Zhejiang University, Zhejiang, China and Shanghai AI laboratory, Shanghai, China (Email: 12321254@zju.edu.cn).}
\thanks{Zuhong Liu is with Shanghai Jiao Tong University and École Polytechnique, Palaiseau, France (Email: zuhong.liu@polytechnique.edu).}
\thanks{Yafei Wang is with Shanghai Jiao Tong University, Shanghai, China (Email: wyfjlu@sjtu.edu.cn).}
\thanks{Siheng Chen is with Shanghai Jiao Tong University and Shanghai AI laboratory, Shanghai, China (Email: sihengc@sjtu.edu.cn).}
\thanks{Yanfeng Wang is with Shanghai Jiao Tong University and Shanghai AI laboratory, Shanghai, China (Email: wangyanfeng622@sjtu.edu.cn).}}



\maketitle

\begin{abstract}
Roadside unit (RSU) can significantly improve the safety and robustness of autonomous vehicles through Vehicle-to-Everything (V2X) communication. Currently, the usage of a single RSU mainly focuses on real-time inference and V2X collaboration, while neglecting the potential value of the high-quality data collected by RSU sensors.  Integrating the vast amounts of data from numerous RSUs can provide a rich source of data for model training. However, the absence of ground truth annotations and the difficulty of transmitting enormous volumes of data are two inevitable barriers to fully exploiting this hidden value. In this paper, we introduce FedRSU, an innovative federated learning framework for self-supervised scene flow estimation. In FedRSU, we present a recurrent self-supervision training paradigm, where for each RSU, the scene flow prediction of points at every timestamp can be supervised by its subsequent future multi-modality observation. Another key component of FedRSU is federated learning,  where multiple devices collaboratively train an ML model while keeping the training data local and private. With the power of the recurrent self-supervised learning paradigm, FL is able to leverage innumerable underutilized data from RSU. To verify the FedRSU framework, we construct a large-scale multi-modality dataset RSU-SF. The dataset consists of 17 RSU clients and an additional 4 vehicle clients, covering various scenarios, modalities, and sensor settings. Based on RSU-SF, we show that FedRSU can greatly improve model performance in ITS and provide a comprehensive benchmark under diverse FL scenarios. To the best of our knowledge, we provide the first real-world LiDAR-camera multi-modal dataset and benchmark for the FL community. Code and dataset are available at \href{https://github.com/wwh0411/FedRSU}{https://github.com/wwh0411/FedRSU}
\end{abstract}


\section{Introduction}


\IEEEPARstart{R}{oadside} units (RSUs) are an important constituent of intelligent transportation systems (ITS).
They are often installed along roadsides, intersections, and other transportation scenarios and outfitted with various sensors and communication devices~\cite{guerna2022roadside, magsino2022enhanced, ackels2021survey}. With the rapid evolution of vehicle-to-infrastructure communication technologies~\cite{chen2017vehicle}, RSUs are able to offer significant added value for intelligent autonomous vehicles~\cite{xu2022v2x, hu2022where2comm, ren2023interruption}. 
At present, the RSU perception system only employs fixed machine-learning models for real-time situation awareness. Generally, these models are pre-trained on limited labeled datasets. Despite the fact that RSUs are equipped with high-resolution sensors capable of providing stable and precise scene observations, the continuously collected streaming data has not been effectively utilized to adapt the model to the current perceptual scene or to improve the model's perceptual capability over time. From the model training perspective, a substantial portion of the streaming data remains underutilized.

\begin{figure}
    \centering
    \includegraphics[width=0.95\linewidth]{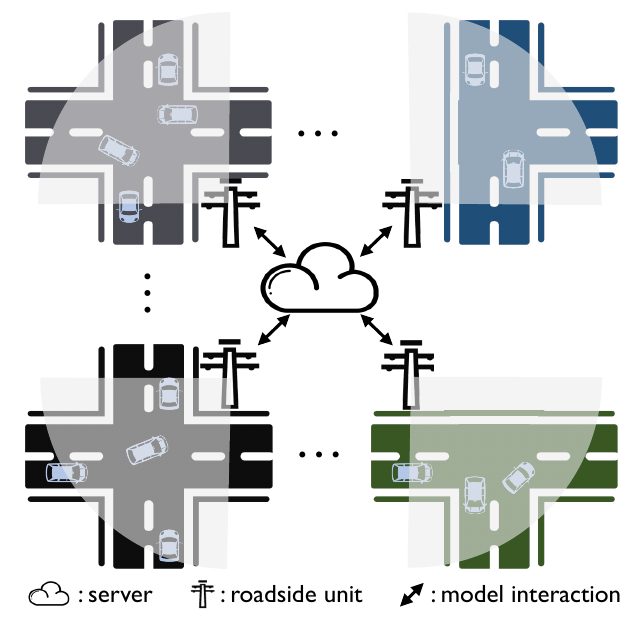}
    \vspace{-3mm}
    \caption{FedRSU system overview, where multiple roadside units (RSUs) collaboratively train a scene flow estimation model without transmitting raw data under the coordination of a cloud server. Iteratively, each RSU trains a local model in a self-supervised manner, and the server aggregates local models. FedRSU can significantly alleviate the challenges of tedious labeling and limited data for one single RSU.}
    \label{fig:enter-label}
    \vspace{-3mm}
\end{figure}

Two formidable challenges need to be addressed in order to leverage the infinite streaming data from RSU sensors to facilitate model training. 
(1) \textbf{Traffic data annotation.} The data labeling process is exceedingly arduous for transportation data. For RSU perception, the annotation of traffic data is crucial and advantageous for model training. However, the annotation process is always arduous and time-consuming, requiring extensive data storage, transmission, and manual labeling. The creation of annotated datasets~\cite{Dair-v2x, busch2022lumpi, wang2022ips300+, Rope3d, A9-I} is always a laborious and expensive endeavor. It is unrealistic to annotate all the data collected by RSUs to continuously refine the perception model. 
(2)\textbf{Data integration.} Integrating data or models from multiple RSU edges is not trivial. In the intelligent transportation system, it is challenging to gather a massive amount of unlabeled data from various sources and train a universal model, as successfully achieved in computer vision~\cite{he2022masked} or natural language processing~\cite{devlin2018bert}. The traffic scene data collected by RSUs may comprise a lot of sensitive information and cannot be shared publicly, such as camera data that contains sensitive facial features and vehicle license plate information. Moreover, the sheer amount of data makes it impossible to aggregate for centralized training due to communication bandwidth and storage limitations.


In our design, we designate scene flow estimation as the principal task for each RSU. Scene flow~\cite{vedula1999three}, which is the 3D equivalent of optical flow~\cite{fortun2015optical} and describes the motion vector of points in 3D space.
Reasoning about motion and forecasting the future of a dynamic scene is particularly critical. Scene flow estimation can be a crucial task that supports various downstream tasks, including motion segmentation~\cite{slim}, object detection~\cite{huang2022representation}, motion prediction~\cite{pillar_motion, fang2024self}, and more. Also, scene flow is a commonly used representation format in self-supervised learning methods~\cite{mittal2020just, pointpwc, Flowstep3d}.

\begin{figure}
    \centering
    \includegraphics[width=0.95\linewidth]{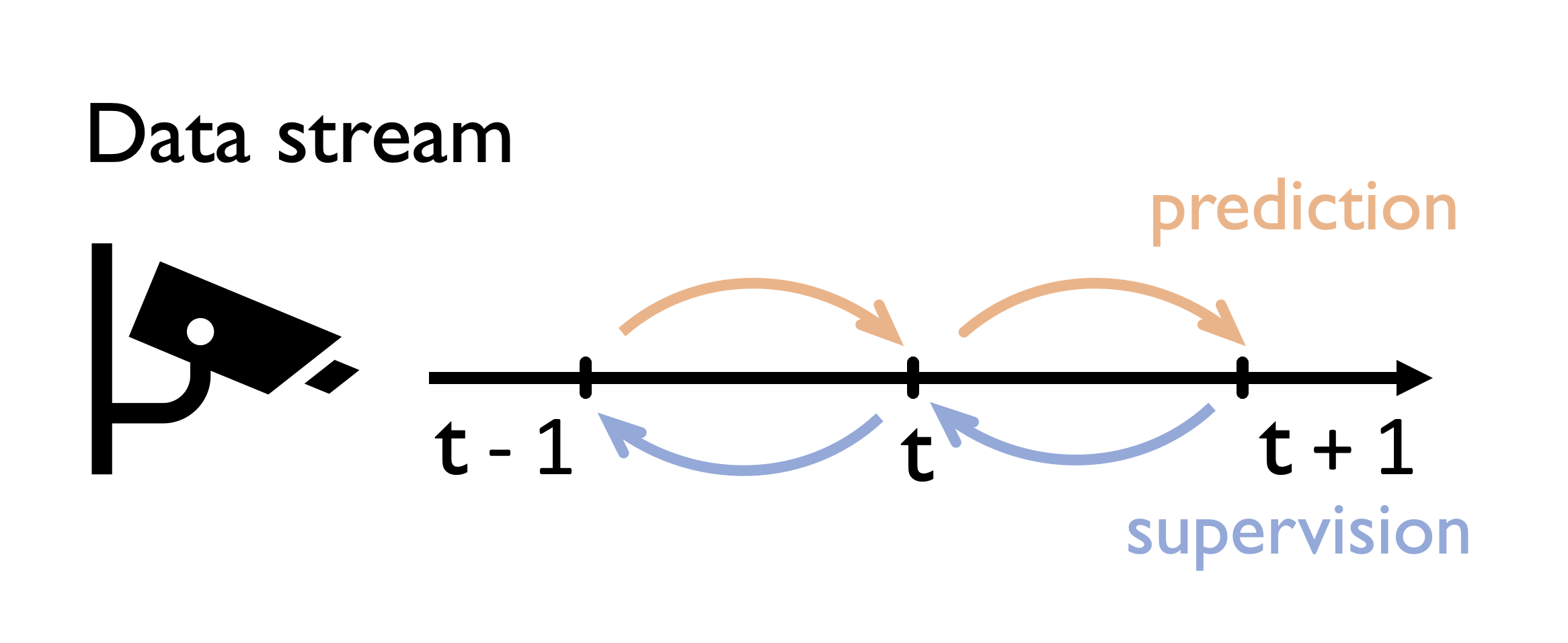}
    \vspace{-3mm}
    \caption{The recurrent self-supervised learning paradigm. The prediction of the model can be supervised by the following frame of sensor data in a self-supervised manner. With the continuous data stream, the model can be continuously improved.}
    \label{fig:recurrent_ssp}
    \vspace{-5mm}
\end{figure}

To address the challenge of \textbf{traffic data annotation}, our core idea is to use a recurrent self-supervision paradigm on every RSU client for scene flow estimation learning, as is shown in Fig.~\ref{fig:recurrent_ssp}. In transportation scenarios, one popular strategy of self-supervised learning is to obtain model guidance from sequential data~\cite{weng2022s2net, khurana2023point, huang2022representation}. With the constant stream of data, real-time scene flow prediction results of the deployed model can be supervised by subsequent future observations from RSU sensors. In this way, the recurrent self-supervised learning paradigm consumes real-time sequential data, allowing for continuous learning and improvement. 


To address the challenge of \textbf{data integration}, we introduce federated learning (FL)~\cite{advances,yang_survey,li_survey} as a promising solution. Federated learning offers a paradigm to collaboratively train a machine learning model under the coordination of a central server while keeping all training data on local devices.
With the power of FL, each RSU equipped with sensors and computational devices can train a local scene flow estimation model using its own extensive data and upload the model parameters for FL aggregation instead of transferring masses of raw data. In the whole training process, all training data are maintained decentralized.
Since RSUs are stationarily deployed at intersections, highways, parking lots, etc., the scene perceived by each RSU is fixed and limited. By fusing information from multiple RSUs in FL, the final global scene flow estimation model will be equipped with higher perception capability and generalization ability than the isolatedly trained local model.


Integrating both solutions mentioned above, in this paper, we propose FedRSU, an FL framework for self-supervised scene flow estimation in the RSU scenarios. The FedRSU framework consists of numerous RSU clients and a centralized server. On each RSU client, a scene flow estimation model is deployed and the model is continuously fine-tuned in a self-supervised manner. 
Additionally, given that RSUs are typically equipped with multi-modal sensors, we propose a multi-model self-supervised method that trains the scene flow estimation model by leveraging information from both point cloud and image.
Besides all RSU clients, a central cloud server coordinates models among all clients. All trained local models on RSU clients are uploaded, aggregated, and combined into a single global model. The global model encapsulates group knowledge and is subsequently distributed back to each client for further refinement. Through multiple rounds of client-server interactions, the resulting global model acquires an enhanced ability to comprehend diverse real-world scenes, a feat that is challenging for a single RSU to accomplish due to its limited perceptible area. 
Consequently, our proposed FedRSU can enable more accurate and reliable perception capabilities in intelligent transportation systems, enhancing the effectiveness and security of the system.

Previous attempts applying federated learning (FL) to ITS have covered various tasks and applications, including perception tasks (e.g., object detection~\cite{rjoub2021improving, wang2022federated, liu2020fedvision}, semantic segmentation~\cite{fantauzzo2022feddrive, shenaj2023learning}, and bird's eye view segmentation~\cite{song2023fedbevt}), prediction tasks (e.g., steering wheel angle prediction~\cite{zhang2021real, zhang2021end} and trajectory prediction~\cite{han2022federated}), and motion control~\cite{zeng2022federated} or driver monitoring applications~\cite{yuan2023federated, zhao2023fedsup}. However, previous efforts have serious problems in their designs and frameworks, whereas our proposed FedRSU boasts three distinct advantages over them. 1) The scene flow estimation task is critical and helpful for 3D scene understanding, which directly benefits RSU perception and overall ITS development. 2) We use RSUs as FL clients instead of onboard devices. Onboard devices have limited computing capability and cannot be employed for model training while carrying out real-time inference tasks in complex and dynamic traffic scenarios. 3) Previous works introduce the assumption of client-side ground truth labels, while raw sensor data generated by traffic participants are unable to be annotated due to privacy and data transmission limitations. Consequently, FL with supervised tasks cannot be practically deployed in a real-world ITS. 
It is important to note that FedRSU features a multi-modality nature. Our proposed dataset includes data from both LiDAR and camera sensors. Additionally, we propose a method for self-supervised learning using multi-modality data. 

Previous RSU datasets~\cite{Dair-v2x, busch2022lumpi, wang2022ips300+, Rope3d, A9-I} only contain data from a single scenario and provide 3D bounding box annotation for scene perception.
To validate our scene flow FL framework FedRSU, we propose a large-scale RSU scene flow dataset, RSU-SF. The RSU-SF dataset comprises a total of 31,311 samples collected from 17 RSUs, with each RSU acting as a client in FL. The RSU scenes are derived from three RSU datasets~\cite{Dair-v2x, busch2022lumpi, wang2022ips300+} and a self-collected dataset, providing a diverse range of scenarios and data heterogeneity. 
To verify and discuss whether the sensor data from vehicles connected to the RSU can effectively participate in the FedRSU system, we selected vehicle-edge data from four different intersections in the DAIR-V2X~\cite{Dair-v2x} dataset as 4 vehicle clients.
Data heterogeneity in the RSU-SF dataset arises from various aspects, including the diverse scenarios in which the RSUs are deployed, the different classes and densities of objects in these scenarios, and the varied modalities, sensor parameters, and sensor settings of the RSUs.

Based on the RSU-SF dataset, we validate the proposed FedRSU from three aspects: 
(1) We conduct experiments under two classical FL settings: generalized FL and personalized FL. For the generalized FL setting, FedRSU improves $33.25\%$ over local learning in terms of epe3d metric and exhibits robust generalization to new clients not seen in the training data. For the personalized FL setting, FedRSU enhances the performance of each participating RSU client.
(2) The proposed multi-modal scene flow learning method on RSU clients facilitates enhanced performance of scene flow models across diverse experimental settings. Additionally, as more clients have multi-modal data, the overall performance of FedRSU models increases monotonously.
(3) We demonstrate that FedRSU is a general system that can incorporate a wide range of existing self-supervised scene flow methods and FL techniques. We benchmarked an assortment of methods within our FedRSU framework. Specifically, we implement three mainstream self-supervised methods for learning scene flow models at the client side~\cite{liu2019flownet3d, pointpwc, Flowstep3d}. Orthogonally, we integrate diverse FL methods, including six generalized FL methods~\cite{fedavg,fedavgm,fedprox,scaffold,feddyn,fednova,fedopt} and seven personalized FL methods~\cite{fedavg,fedprox,ditto,fedper,fedrep,pfedme,pfedgraph}. This compatibility indicates the significant potential for further improvement of FedRSU in tandem with ongoing advancements in both the fields of self-supervised scene flow learning and federated learning.

Overall, the key contributions of this work are as follows:
\begin{itemize}
    \item We propose a new and practical federated learning framework on roadside units (FedRSU), where multiple RSUs collaboratively train a scene flow estimation model in a self-supervised manner.
    \item We propose a novel multi-modal scene flow learning method on each RSU client, which leverages image data to guide scene flow learning.
    \item We construct a diverse and practical scene flow dataset RSU-SF to promote the development of FedRSU and FL.
    \item We conduct extensive experiments on multiple baselines and scenarios to provide more insights and call for more future explorations.
\end{itemize}

\textbf{Outline}. This paper is structured as follows:
In Section~\ref{sec:related_work}, we introduce related works. 
In Section~\ref{sec:fedrsu}, we formulate the proposed setting, introduce the FedRSU framework, and our proposed federated multi-modal self-supervised learning algorithm. 
In Section~\ref{sec:dataset}, we introduce the constructed dataset RSU-SF for scene flow estimation and federated learning. 
In Section~\ref{sec:exp}, we conduct extensive experiments on diverse baselines and scenarios.
In Section~\ref{sec:discuss}, we provide discussions on future directions and limitations.
In Section~\ref{sec:conclude}, we summarize the paper.

\section{Related Work}
\label{sec:related_work}

\subsection{Roadside Units in Intelligent Transportation Systems}
\label{sec:rsu_related_work}



The domain of RSU perception in ITS requires specialized data and methods, given that the data format for RSU perception differs significantly from that of general autonomous driving datasets~\cite{nuscenes, waymoopen}. The sensors in autonomous driving datasets are typically mounted relatively close to the ground, whereas RSUs provide an aerial perspective with a broader field of view and fewer occlusions. Consequently, numerous datasets~\cite{Rope3d, busch2022lumpi, wang2022ips300+, A9-I} have been proposed for this field, and many perception methods~\cite{BEVHeight, hu2023aerial, zimmer2023infradet3d} have been developed to cater to these specific scenarios. \cite{BEVHeight} predicts the height of the 3D object to enhance monocular 3d detection. \cite{zimmer2023infradet3d} focuses on the camera-lidar fusion problem for RSU.
Besides, a large number of studies have explored the collaboration setting~\cite{Dair-v2x, V2X-Seq, V2X-Sim} and algorithms~\cite{arnold2020cooperative, hu2022where2comm} between RSUs and autonomous vehicles.

Due to the singular nature of RSU scenarios and the high cost of annotation labeling, obtaining sufficient training data for RSU remains a critical challenge.
In existing RSU datasets, the RSU data is generally gathered from a limited number of locations, resulting in low environmental diversity and weak generalization capability of trained models. Additionally, obtaining large amounts of labels incurs significant expenses.
To tackle this issue, ~\cite{zhang2023robust} proposed utilizing Augmented Reality~\cite{azuma1997survey} and Generative Adversarial Network~\cite{creswell2018generative} to produce synthesized data. ~\cite{wu2023efficient} design a semi-automated scheme for labeling RSU data. 

In this paper, we attempt to fundamentally address this issue from a system design by introducing federated learning and a self-supervision paradigm.

\subsection{Scene Flow on Point Clouds}
Scene flow, representing the 3d motion of points in space, is first introduced in \cite{vedula1999three}. Following ~\cite{vedula1999three}, a series of methods estimate scene flow under the setting of stereo camera setting~\cite{huguet2007variational, carceroni2002multi, pons2005modelling, vogel20153d}. Additionally, scene flow can also be estimated from RGB-D image\cite{brickwedde2019mono, rishav2020deeplidarflow, shao2018motion, teed2021raft}. With the advancement of LiDAR applications and deep learning techniques for point clouds~\cite{pointnet, pointnet++, lang2019pointpillars}, recent works tend to directly estimate scene flow from pairs of LiDAR point clouds. 
From FlowNet3D~\cite{liu2019flownet3d}, which is built upon pointnet++\cite{pointnet++}, different network architectures and losses~\cite{puy2020flot, gu2019hplflownet, fastflow3d, cheng2022bi, li2021hcrf} are proposed to improve the estimation of 3d scene flow on point clouds.

Meanwhile, many point-cloud-based methods~\cite{pointpwc, mittal2020just, slim, Flowstep3d, li2022rigidflow, tishchenko2020self} explore learning scene flow in a self-supervised manner.
\cite{pointpwc} proposes employing a combination of chamfer distance, smoothness constraint, and Laplacian regularization\cite{sorkine2005laplacian} to guide scene flow training
\cite{li2022rigidflow, dong2022exploiting} utilize ego-motion estimation and the piecewise rigid nature of point clouds.
\cite{Flowstep3d} proposes to estimate the scene flow iteratively using a recurrent architecture and the same losses as \cite{pointpwc}.

In this paper, we choose several state-of-the-art self-supervised scene flow methods~\cite{mittal2020just, pointpwc, Flowstep3d} as the basis of the federated learning framework.


\subsection{Federated Learning on Data Heterogeneity}

Data heterogeneity is one key challenge in FL~\cite{advances,yang_survey}, which is shown to bring adverse effects empirically~\cite{fedavg,fed_empirical} and theoretically~\cite{li2019convergence,wang2021cooperative}. Addressing this issue, a series of algorithms~\cite{fedprox,scaffold}, datasets~\cite{flair,flamby}, and benchmarks~\cite{leaf,pflbench} are proposed.

\textbf{Algorithms.} Generally, FL can be divided into two sub-fields: generalized FL (gFL) and personalized FL (pFL), where gFL aims for training a generalized global model and pFL aims for training multiple personalized local models~\cite{advances,li_survey}. 1) Targeting gFL, many methods are proposed based on model regularization~\cite{fedprox,feddyn}, feature regularization~\cite{moon,fedfm}, control variate~\cite{scaffold,vrlsgd}, model momentum~\cite{fedavgm,fedopt}, knowledge distillation~\cite{feddf,fedgen} and aggregation weight adjustment~\cite{fednova,feddisco}. 2) Targeting pFL, many methods are proposed from the perspective of model regularization~\cite{ditto,pfedme}, meta learning~\cite{per_ml_1,per_ml_2}, aggregation~\cite{pfedgraph,cfl}, and model partitioning~\cite{fedper,fedrep}. In our benchmark, we conduct extensive performance evaluations on multiple representative gFL and pFL methods, serving as an experimental study to provide more insights for future research.

\textbf{Datasets.} Many FL papers simulate data heterogeneity by artificially partitioning existing classic datasets for 1) category heterogeneity~\cite{fedavg,fedma,bayesian}, which includes MNIST~\cite{mnist}, Fashion-MNIST~\cite{xiao2017fashion}, CIFAR-10/100~\cite{cifar10}, CINIC-10~\cite{darlow2018cinic} or ImageNet~\cite{imagenet}; and 2) domain heterogeneity~\cite{fed_da,fedbn,fedsr}, which includes Digits~\cite{mnist,svhn}, Office-Home~\cite{office-home} and DomainNet~\cite{domainnet}. However, such synthetic data partition may fail to well model the real-world federated data distributions~\cite{what}. Thus, some start to consider more realistic datasets that are partitioned by user identifier, which include Sentiment140~\cite{sentiment140}, iNaturalist~\cite{inaturalist}, Landmarks~\cite{landmark}, FLamby~\cite{flamby} and FLAIR~\cite{flair}. 

\textbf{Benchmarks.} To promote fair comparison and re-productivity, some benchmarks in FL are proposed. LEAF~\cite{leaf} releases specific user partitions on six text and image classification datasets. FedML~\cite{fedml} and FedScale~\cite{fedscale} provide systems for multiple tasks while FedNLP~\cite{fednlp} and FederatedScope-GNN~\cite{federatedscope} focus on NLP and graph, respectively. pFL-Bench~\cite{pflbench} specifically benchmarks personalized FL. 

However, all these datasets and benchmarks are targeted for supervised tasks with only one modality. In contrast, FedRSU demonstrates a brand new practical scenario, which is a realistic and multimodal dataset for self-supervised tasks. Besides, we provide comprehensive benchmarks and experimental studies for both generalized and personalized FL.

\section{FedRSU: Federated Learning on RSUs}
\label{sec:fedrsu}

In this section, we formulate the problem of scene flow estimation, describe our novel design of multi-modal recurrent self-supervised learning for scene flow estimation at each RSU, and finally present the overall FedRSU system.

\begin{figure*}
    \centering
    \includegraphics[width=0.85\textwidth]{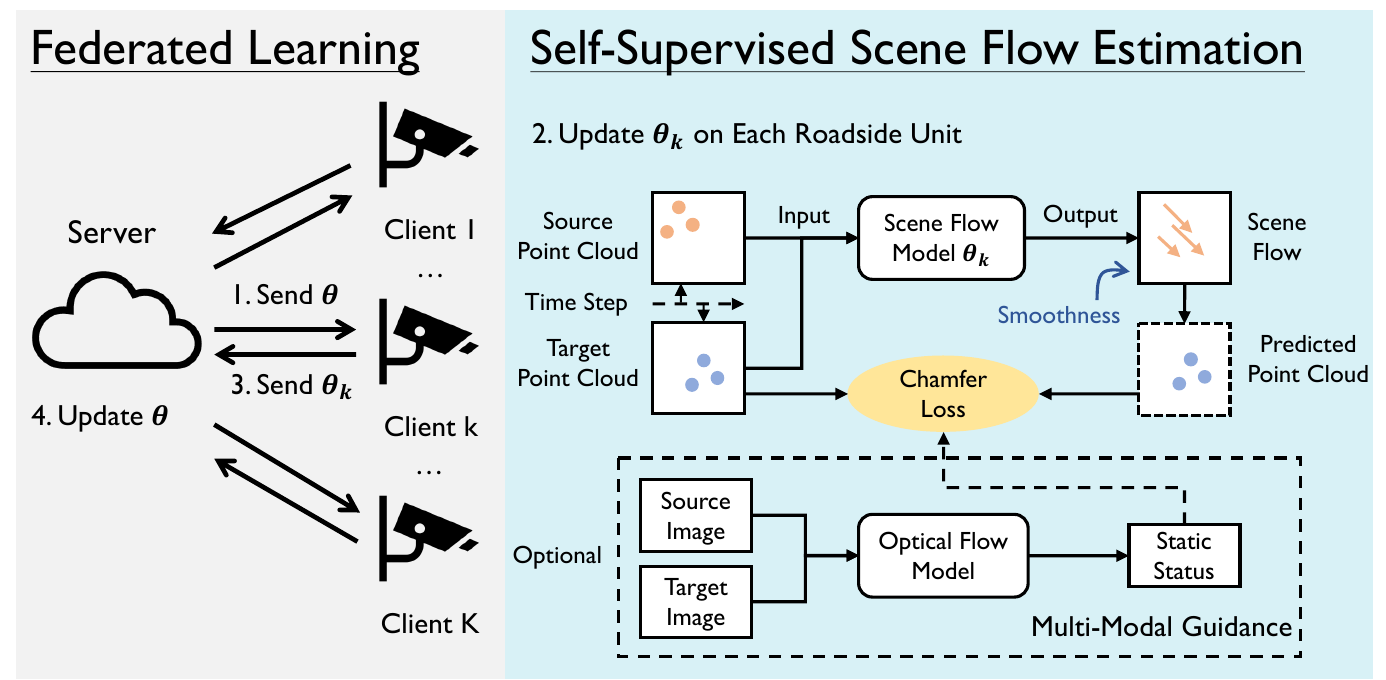}
    \vspace{-3mm}
    \caption{Overview of FedRSU framework.  FedRSU consists of four steps. 1) The server sends the global model to all available clients, 2) each client updates local model supervised by Chamfer loss and smoothness regularization, 3) each client sends local model to the server, 4) the server updates global model by aggregating received local models. These four steps will iterate for multiple rounds.}
    \label{fig:fedrsu}
    \vspace{-3mm}
\end{figure*}

\subsection{Problem Formulation}

The focused task is scene flow estimation that describes the motion vector of points in 3D space, which is a crucial component to support various downstream tasks, including segmentation~\cite{slim}, instance segmentation~\cite{song2022ogc}, object detection~\cite{huang2022representation}, motion prediction~\cite{pillar_motion}, trajectory prediction~\cite{najibi2022motion}, and more.
To achieve this, our core goal is to train a scene flow estimation model on the constant stream of RSU data in a recurrent self-supervised paradigm. As is shown in Fig.~\ref{fig:recurrent_ssp}, in the data stream of RSU sensors, the prediction at each frame can be supervised by its following future frame. Therefore, in our method, the denotation $t$ can be any frame in the data stream. 

Denote the dataset as $\mathcal{D}=\{ (\mathbf{X}^{(pc)}_i, \mathbf{X}^{(img)}_i)\}_{i=1}^{N}$, where $N$ is the number of samples of the dataset.
$\mathbf{X}^{(pc)}_i=(\mathbf{P}^{t-1}_i, \mathbf{P}^{t}_i)$, where source point cloud $\mathbf{P}^{t-1}_i = \left\{ p^{t-1}_a \in \mathbb{R}^3 \right\}_{a=1}^{n_1}$ and target point cloud $\mathbf{P}^{t}_i = \left\{ p^{t}_b \in \mathbb{R}^3 \right\}_{b=1}^{n_2}$ are from two consecutive time frames.
$\mathbf{X}^{(pc)}_i=(\mathbf{I}^{t-1}_i, \mathbf{I}^{t}_i)$ are the corresponding images.


Basically, the objective of scene flow estimation is to estimate a motion vector $f_a \in \mathbb{R}^3$ of point $p^{t-1}_a \in \mathbb{R}^3$ from the first frame $\mathbf{P}^{t-1}_i$ to its possible new position in the second frame $\mathbf{P}^{t}_i$.
Due to the data sparsity of LiDAR point clouds and occlusion caused by moving objects, $p_a$ may not have its corresponding point in $\mathbf{P}^{t}_i$ and the point numbers $n_1$ and $n_2$ may differ.
Therefore, the predicted flow $\mathbf{F}_i = \left\{ f_a \in \mathbb{R}^3 \right\}_{a=1}^{n_1}$ is not the point-to-point correspondences between $\mathbf{P}^{t-1}_i$ and $\mathbf{P}^{t}_i$, but the motion representation describing the scene.

\subsection{Scene Flow Estimation via Multi-modal Recurrent Self-Supervised Learning}
\label{sec:method_self}

For each RSU, it trains a local scene flow estimation model on its locally perceived data. 
Considering the impracticality of annotating RSU's data (which is arduous and time-consuming) and the limitation of scene understanding from a single modality, we propose to train the flow estimation model in a multi-modal recurrent self-supervised manner.

There are two key designs in our framework.
1) Recurrent self-supervision: based on the continuously coming sensor data, each RSU is trained to predict the scene flow between two sequential scene frames, thereby constructing supervision from the data itself rather than by human annotating.
Such naturally existent supervision signal enables us to train the scene flow estimation model without any human labeling.
2) Multi-modality: based on the synchronously captured image data from cameras and point cloud data from LiDARs, we propose to leverage them during self-supervised learning. Therefore, these two modalities can complement each other to improve the scene flow estimation.

\begin{figure}
    \centering
    \includegraphics[width=0.95\linewidth]{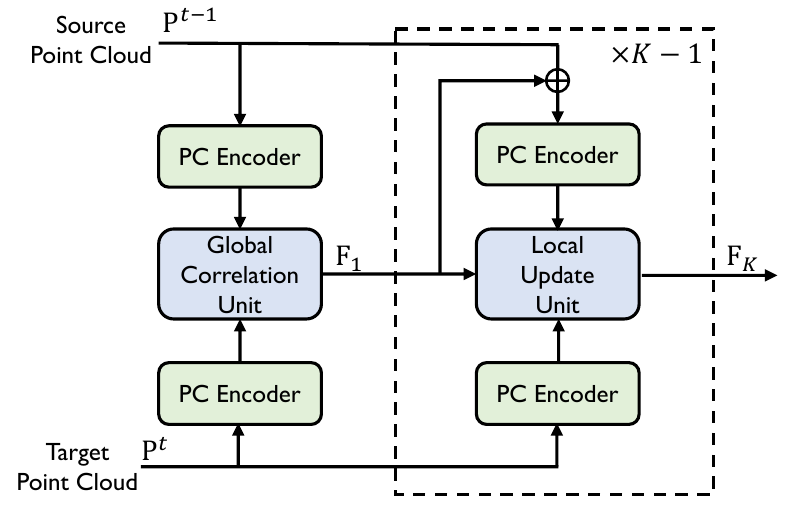}
    \vspace{-3mm}
    \caption{Overview of the scene flow model architecture. We follow the architecture of Flowstep3d~\cite{Flowstep3d} and predict the scene flow in a coarse to fine manner.}
    \label{fig:model_arch}
    \vspace{-3mm}
\end{figure}

\textbf{Model architecture.} For the scene flow model, we follow the recurrent model architecture in Flowstep3D~\cite{Flowstep3d} which predicts the scene flow from coarse to fine progressively. An overview is shown in Fig.~\ref{fig:model_arch}
The model predicts a sequence of flow $\{ \mathbf{F}_k \in \mathbb{R}^{n_1 \times 3} \}_{k=1}^{K}$, where $\mathbf{F}_K$ is the final scene flow estimation output. 

The architecture of the scene flow estimation model is mainly composed of three modules: point cloud encoder, global correlation unit, and local update unit. The point cloud encoder, following \cite{liu2019flownet3d}, extracts the features of raw point clouds. For the first round, in global correlation unit, a correlation encoder is used to capture the relationship between two point clouds and predict the coarse scene flow $\mathbf{F}_1$. Following \cite{puy2020flot}, the correlation encoder calculates the cosine similarity between each pair of point features and constructs an all-to-all correlation matrix. 
From the second round, the local update unit makes a refinement on flow estimation based on predicted results from previous iterations. 
For iteration $k$, a flow estimator suggested by \cite{liu2019flownet3d} predicts the scene flow $\Delta \mathbf{F}_{k-1}$ from $\mathbf{P}^{t-1\prime}$ to $\mathbf{P}^{t}$, where $\mathbf{P}^{t-1\prime} = \mathbf{P}^{t-1} + \mathbf{F}_{k-1}$. And the predicted scene flow is updated by $\mathbf{F}_{k} = \mathbf{F}_{k-1} + \Delta \mathbf{F}_{k-1}$.

Note that our main technical design is orthogonal to the model architecture and there are multiple architecture candidates~\cite{liu2019flownet3d,pointpwc,Flowstep3d}. Among these, we adopt the FlowStep3D~\cite{Flowstep3d} as our model architecture as it tends to achieve better performance.

\textbf{Multi-modal self-supervised loss.}
Learning scene flow estimation without human labeling calls for self-supervised loss designing such that the model can learn from the data itself. 
To achieve this, we consider two classical self-supervised loss terms, namely, Chamfer distance loss and smoothness regularization loss~\cite{Flowstep3d}.
These two terms encourage local model to predict the accurate flow while preserving appropriate local neighbor structures.

We give a detailed illustration of the implemented multi-modal self-supervised loss on a data pair $(\mathbf{X}^{(pc)},\mathbf{X}^{(img)})$, where we omit the sample subscript for simplicity.
Here, $\mathbf{X}^{(pc)}=(\mathbf{P}^{t-1},\mathbf{P}^{t})$ are two consecutive point clouds from the sensor data stream and $\mathbf{X}^{(img)}=(\mathbf{I}^{t-1},\mathbf{I}^{t})$ are the corresponding image data.
Given the source point cloud $\mathbf{P}^{t-1} = \left\{ p^{t-1}_a \in \mathbb{R}^3 \right\}_{a=1}^{n_1}$ and the target point cloud $\mathbf{P}^{t} = \left\{ p^{t}_b \in \mathbb{R}^3 \right\}_{b=1}^{n_2}$, the scene flow estimation model $h(\bm{\theta};\mathbf{P}^{t-1},\mathbf{P}^{t})$ makes the flow estimation $\mathbf{F}$, where $\bm{\theta}$ is the local model of some client.
The predicted flow consists of $n_1$ motion vectors $f_a$:
$\mathbf{F} = \left\{ f_a \in \mathbb{R}^3 \right\}_{a=1}^{n_1}$.
Then, the predicted point cloud can be denoted as
\begin{equation*}
    \tilde{\mathbf{P}} = \left\{ \tilde{p}_a \in \mathbb{R}^3 \mid \tilde{p}_a = p_a + f_a \right\}_{a=1}^{n_1}.
\end{equation*}

With this predicted point cloud, one basic loss candidate is Chamfer loss, which enforces the source point cloud to move toward the target point cloud based on point cloud only.
However, despite point cloud can offer rich 3D structural information, it has limitations such as lack of color information, ambiguities in object boundaries, noise and artifacts, which can hinder the accurate prediction of scene flow.

To tackle these limitations, we propose to leverage the image information of the corresponding camera to assist self-supervised learning. 
We encode the image information into optical flow, which is utilized to guide the fine-grained Chamfer loss.
In this way, the image data serves to complement the limitations of point cloud data as it provides additional information about color, boundary and mitigates the effects of noise of point cloud. 
Note that the optical flow is only used to assist training, leaving the pipeline during inference the same.

Since the cameras in the RSU are fixed, the background scene and static objects within the perception range are invariant in the image. 
As a result, the optical flow information from camera videos can be acquired easily. 
We exclusively leverage the pre-trained model~\cite{teed2020raft} provided by PyTorch~\cite{paszke2019pytorch} to obtain the optical flow $\mathbf{I}^{opt}$ from $\mathbf{I}^{t-1}$ to $\mathbf{I}^{t}$.

For each point $p_a$, using the transformation information between point cloud coordinates and image space, we can get its corresponding pixel $(u_a, v_a)$. Then the corresponding optical flow of each point can be retrieved.
\begin{equation*}
    f^{opt}_{a} = \mathbf{I}_a^{opt} (u_a, v_a)
\end{equation*}

Then, the probability of the $a$-th point being static~\cite{pillar_motion} is:
\begin{equation*}
    s_a = \exp ( -\alpha \left\| f^{opt}_{a} \right\| ).
\end{equation*}

This probability is subsequently used to define a multi-modal Chamfer loss:
\begin{equation}
\label{eq:cham}
    \mathcal{L}_{ch}(\bm{\theta}) = \sum_{\tilde{p}_a \in \tilde{\mathbf{P}}} s_a \min_{p_b \in \mathbf{P}^{t}} \left\| \tilde{p}_a - p_b \right\|_2^2,
\end{equation}
which assigns larger punishments for those points that are more likely to be static. In this way, the optical flow (image data) can assist the self-supervised learning on point cloud data providing more fine-grained supervision guidance.

Besides, we also apply a smoothness regularization to avoid being stuck in local minima~\cite{Flowstep3d,pointpwc} and preserve local neighbor structures.
Using $\ell_2$ distance, the smoothness regularization loss is defined as:
\begin{equation}
    \mathcal{L}_{reg}(\bm{\theta}) = \sum_{p_a \in \mathbf{P}} \frac{1}{\left| \mathcal{N}(p_a) \right|} \sum_{p_k \in \mathcal{N}(p_a)} \left\| f_a - f_k \right\|_2^2.
\end{equation}

Overall, the multi-modal self-supervised learning loss of each local client is then:
\begin{equation}
\label{eq:local_loss_2}
    \mathcal{L}(\bm{\theta}) = \beta_{ch} \mathcal{L}_{ch} + \beta_{reg} \mathcal{L}_{reg},
\end{equation}
where $\beta_{ch}$ and $\beta_{reg}$ are the hyper-parameters to balance between Chamfer loss and regularization loss. Note that we define the above sample-level loss for simplicity of notations.

\subsection{Federated Learning System}

Though each RSU can obtain a scene flow estimation model following the above procedure, its capability is strongly restrained by the limited perceived scene.
Sharing data among RSUs (Central Learning), as the most direct solution, is faced with two critical practical issues: 
1) the massive continuously generated data would bring too much burden for communication and memory; 
2) recording and sharing data could raise privacy concerns as the raw data usually contains private information.

Addressing this, we propose to a new collaborative learning system FedRSU, which enables collaborative training of scene flow estimation model among multiple RSUs.
Through FedRSU, the final model is essentially the union of knowledge captured by multiple RSUs, breaking the perceptual limits of each individual RSU, thus facilitating more accurate and robust estimation.
Following the conventional procedures of federated learning~\cite{fedavg}, the FedRSU consists of four iterative steps for each communication round, namely, global model broadcasting, local model training, local model uploading, and global model updating. The overview of the FedRSU system is shown in Fig.~\ref{fig:fedrsu}.

\textbf{Global model broadcasting.} In this step, the global model is broadcast to clients to serve for local model initialization. Specifically, at the beginning of each round $t$, the server broadcasts the global model $\bm{\theta}^t$ to each available client $k \in \mathcal{S}^t$, where $\mathcal{S}^t$ consists of an index of all available clients at round $t$. Then, each client $k$ uses this global model to initialize its local model $\bm{\theta}_k^{(t,0)} := \bm{\theta}^t$, where $\bm{\theta}_k^{(t,0)}$ denotes the local model at $t$-th round and $0$-th training iteration.

\textbf{Local model training.} In this step, each client uses its local dataset to train a local model for scene flow estimation, which is guided by the self-supervised loss terms illustrated in Section~\ref{sec:method_self}.
Specifically, starting from the initial local model $\bm{\theta}_k^{(t,0)}$, client $k$ will conduct multiple iterations of SGD updates based on the local dataset $\mathcal{D}_k$. At each iteration $r$, the local model update is represented as:
\begin{equation}
\label{eq:local}
    \bm{\theta}_k^{(t,r+1)} = \bm{\theta}_k^{(t,r)} - \eta \nabla \mathcal{L}(\bm{\theta}_k^{(t,r)},\xi_k),
\end{equation}
where $\mathcal{L}(\bm{\theta}_k^{(t,r)},\xi_k)$ denotes the self-supervised loss computed based on model $\bm{\theta}_k^{(t,r)}$ and a batch $\xi_k$ sampled from dataset $\mathcal{D}_k$. The detailed self-supervised loss function for each data sample is shown in~\eqref{eq:local_loss_2}. After $\tau_k$ iterations, the final trained local model is denoted as $\bm{\theta}_k^{(t,\tau_k)}$.


\textbf{Local model uploading.} In this step, each client uploads its updated local model to the server to serve for aggregating local models. Specifically, each available client $k \in \mathcal{S}^t$ uploads local model $\bm{\theta}_k^{(t,\tau_k)}$ to the server.

\textbf{Global model updating.} In this step, the server updates global model by aggregating local models, which is subsequently broadcast to available clients for the next round. Specifically, the global model is weighted and updated as:
\begin{equation}
    \bm{\theta}^{t+1} := \frac{N_k}{\sum_{i\in\mathcal{S}^t} N_i} \bm{\theta}_k^{(t,\tau_k)}.
\end{equation}

The above four steps iterate for $T$ rounds in total and the system outputs global model $\bm{\theta}^{T}$ at the end.

\section{RSU-SF Dataset}
\label{sec:dataset}

To validate our proposed FedRSU framework, we have established a large-scale RSU scene flow dataset, RSU-SF, consisting of multi-modality sensor data with a high degree of data heterogeneity. In this section, we will provide an overview of the dataset collection process and present relevant dataset statistics.

\subsection{Dataset Collection}

\textbf{Training data formation.} For self-supervised scene flow learning, the training data only requires two sequential frames of LiDAR point clouds. Besides, the corresponding camera image can serve as a supplementary aid for model training. In RSU-SF, the temporal interval between two consecutive frames is set to 0.1 seconds (i.e. a sensor sampling frequency of 10 Hz). Furthermore, we remove the ground points of each scene using a height threshold to simplify the scene flow learning.  In a real-world RSU scenario, it is easy and straightforward to filter out background points since all sensors are deployed stationarily.

\textbf{Scene flow label generation.} To validate and test the models, ground truth scene flow labels are necessary to compute and compare metrics. In RSU-SF, these labels are derived from the 3D bounding boxes and tracking labels of scene participants. The method used to generate ground truth is similar to that employed by previous methods~\cite{slim}. Specifically, we extract the rigid transformation of an object from its consecutive bounding box annotations, and then the flow of points within the bounding box can be directly calculated according to the rigid transformation.

\begin{figure*}[t]
\centering
{\includegraphics[width=0.24\textwidth]{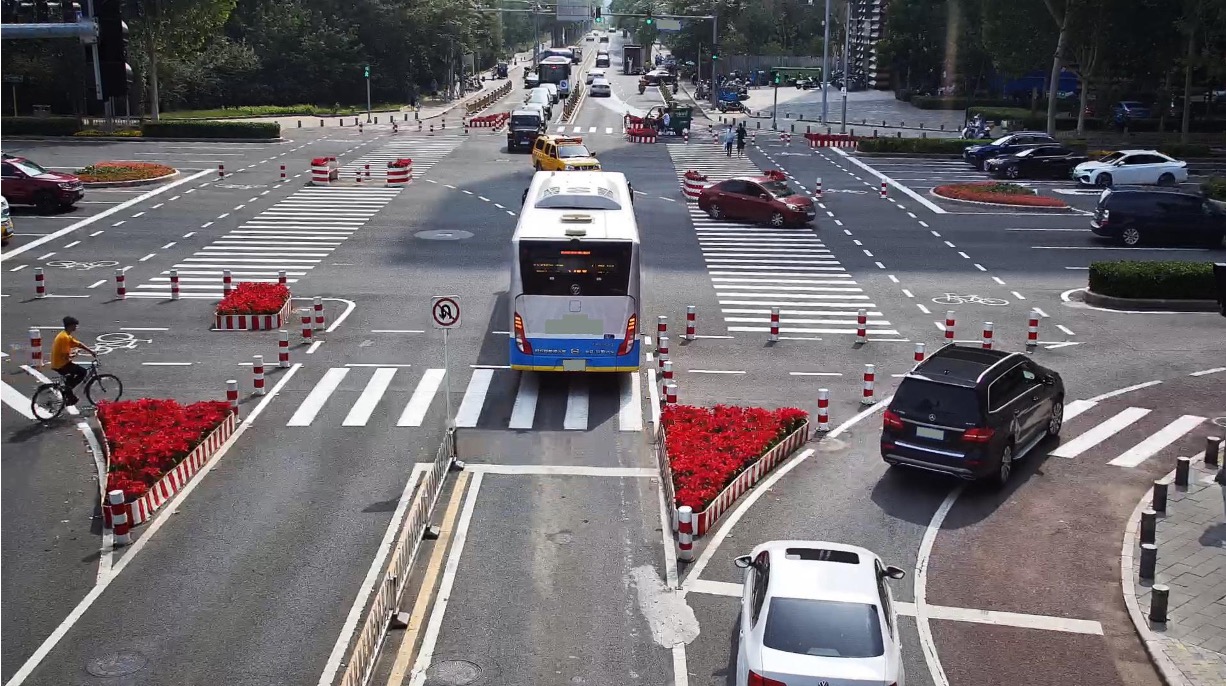}%
\label{fig:dairv2x_img}}
\hfil
{\includegraphics[width=0.24\textwidth]{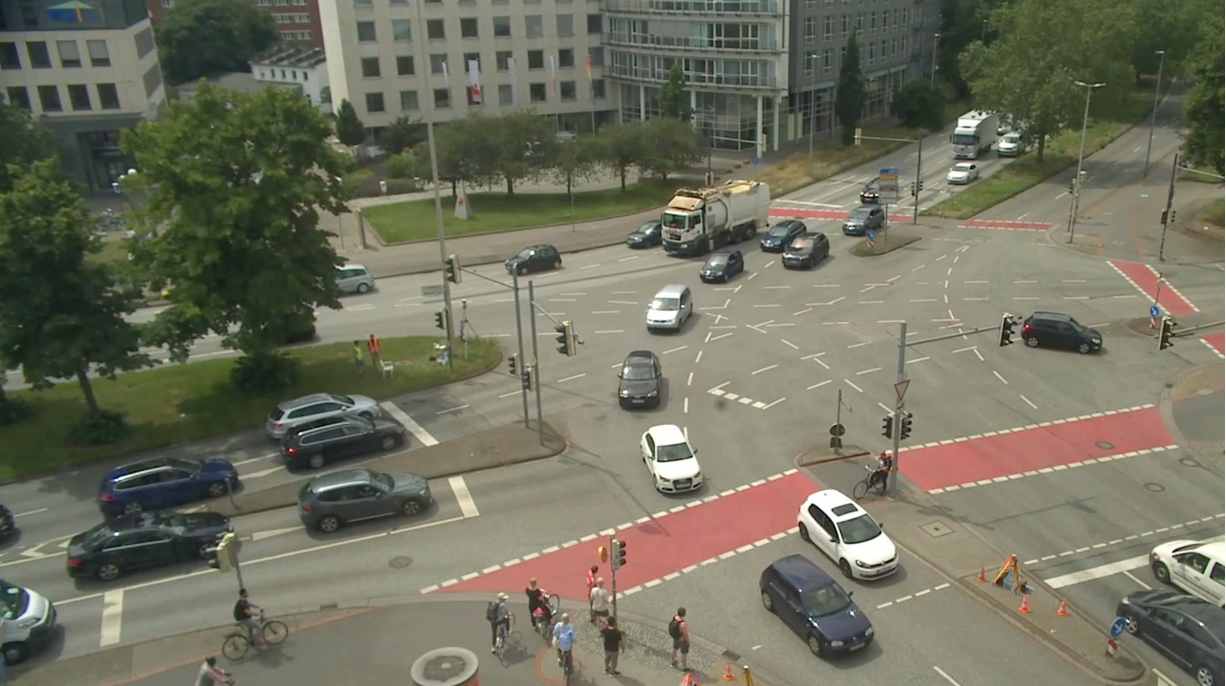}%
\label{fig:lumpi_img}}
\hfil
{\includegraphics[width=0.24\textwidth]{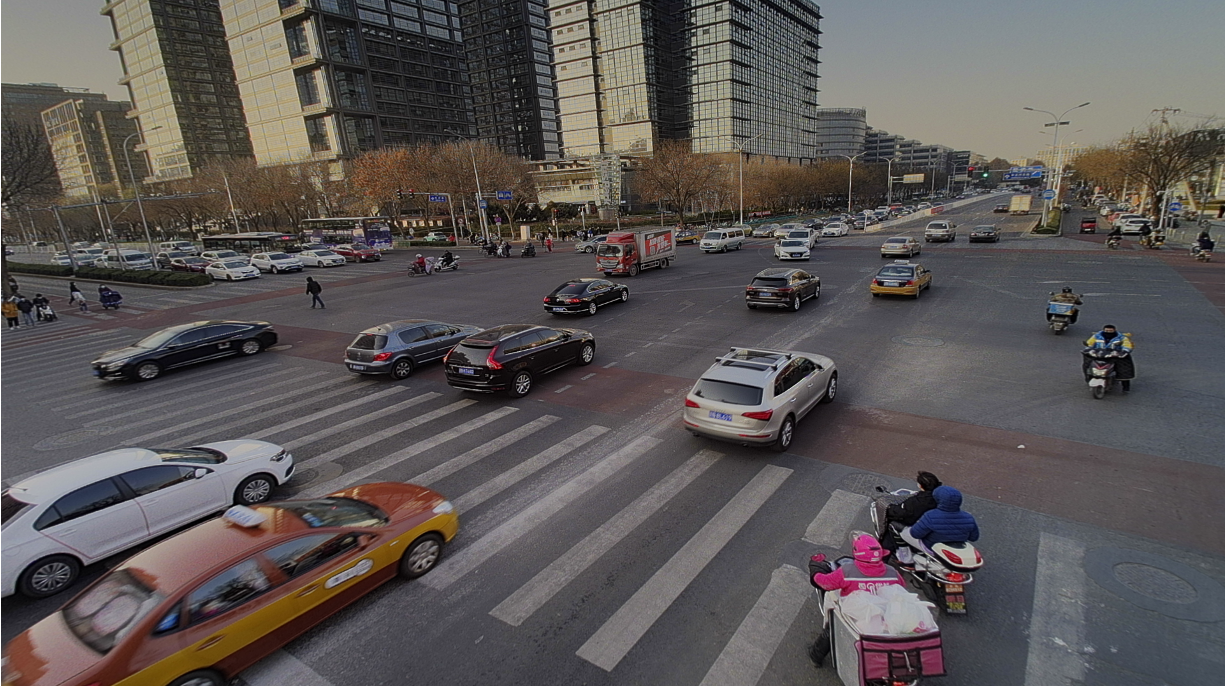}%
\label{fig:ips_img}}
\hfil
{\includegraphics[width=0.24\textwidth]{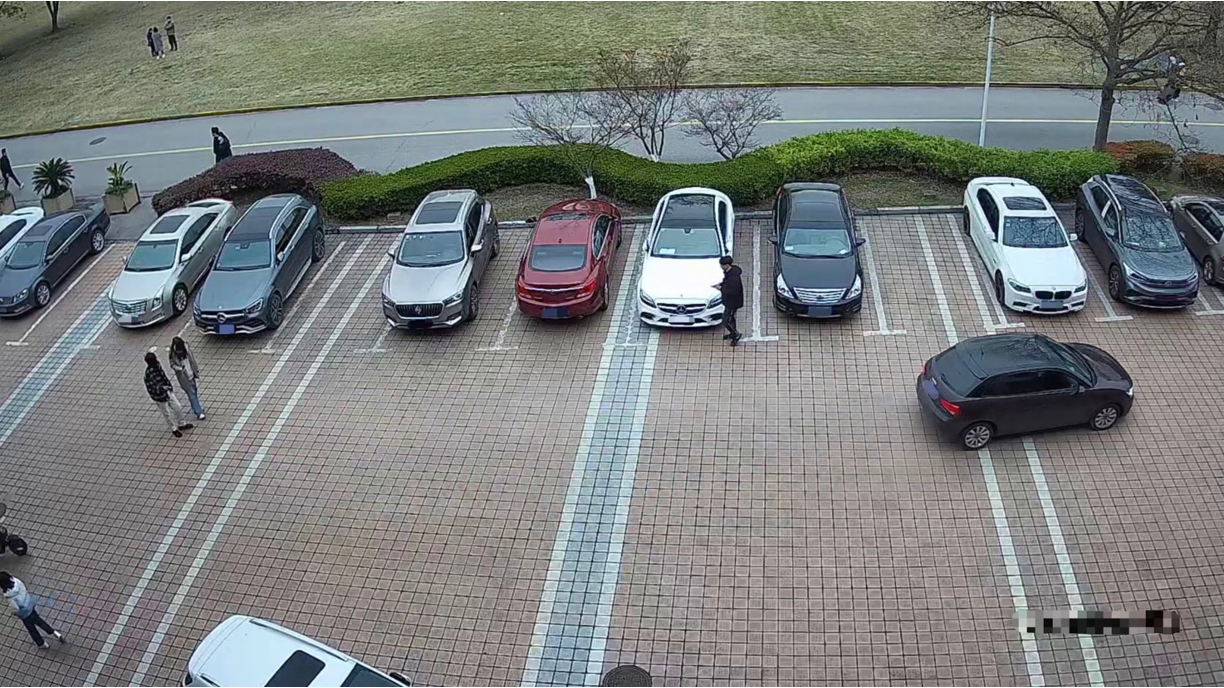}%
\label{fig:campus_img}}
\hfil
\subfloat[DAIR-V2X]{\includegraphics[width=0.24\textwidth]{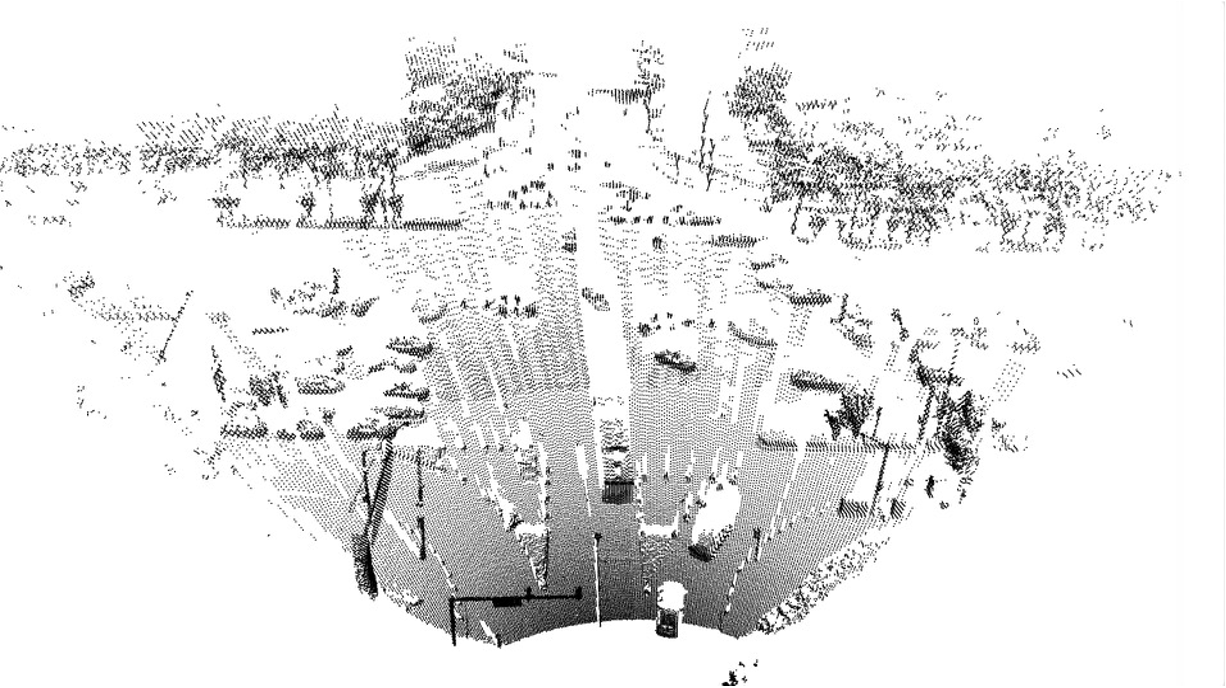}%
\label{fig:diarv2x_pc}}
\hfil
\subfloat[LUMPI]{\includegraphics[width=0.24\textwidth]{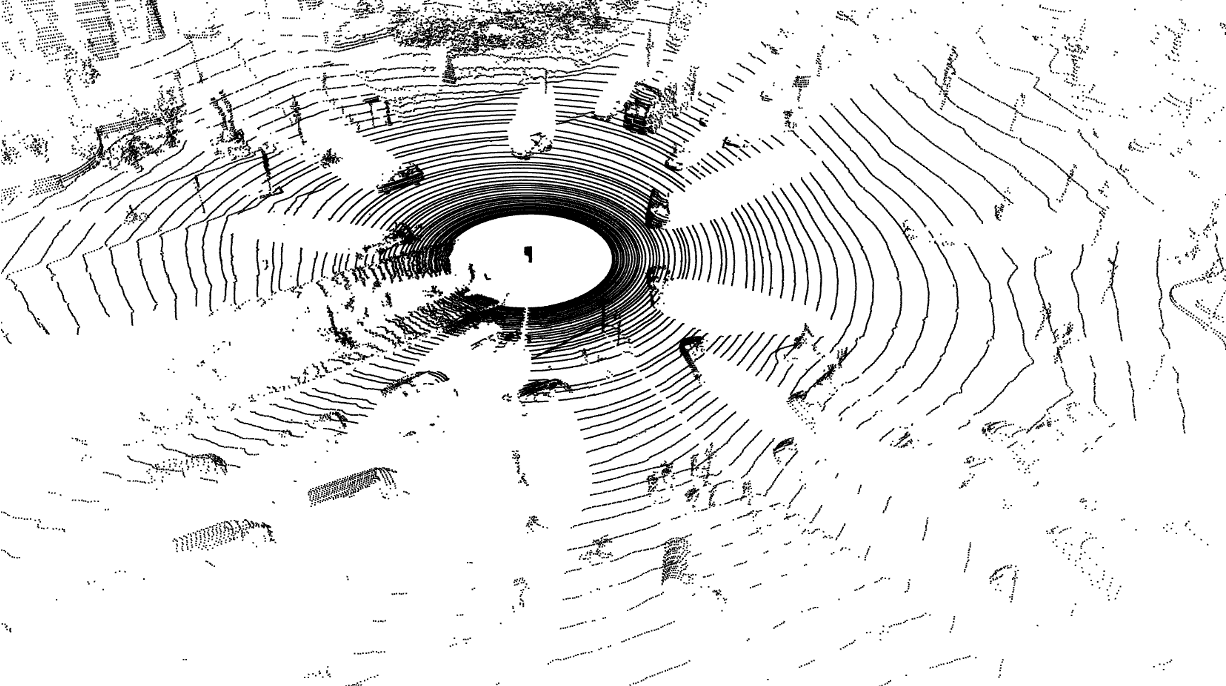}%
\label{fig:lumpi_pc}}
\hfil
\subfloat[IPS300+]{\includegraphics[width=0.24\textwidth]{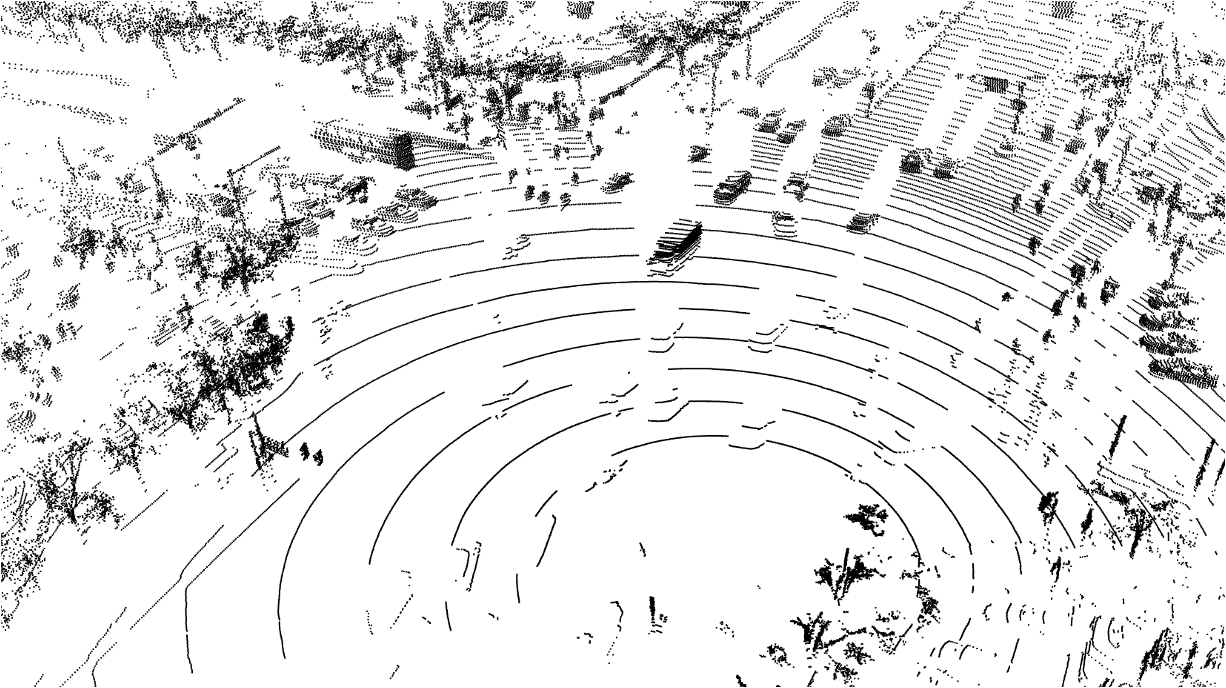}%
\label{fig:ips_pc}}
\hfil
\subfloat[Campus]{\includegraphics[width=0.24\textwidth]{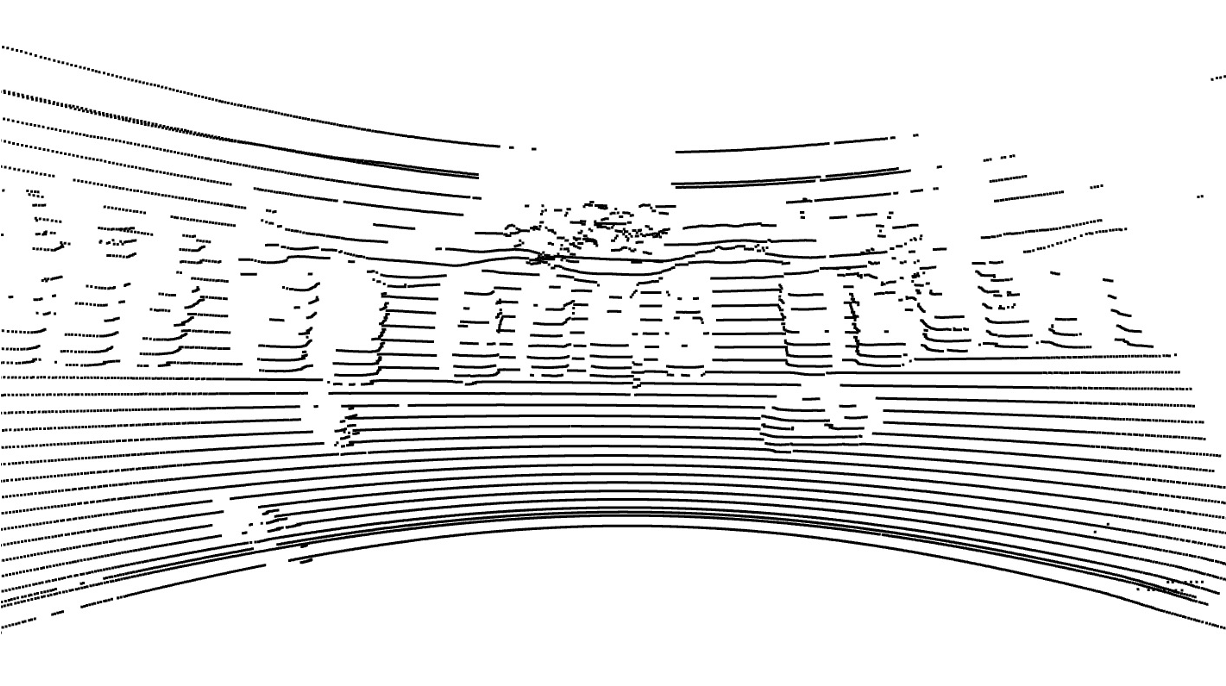}%
\label{fig:campus}}
\caption{Various RSU settings in four base datasets: DAIR-V2X~\cite{Dair-v2x}, LUMPI~\cite{busch2022lumpi}, IPS300+~\cite{wang2022ips300+}, and our collected campus dataset. DAIR-V2X collects data from normal traffic crossroads, LUMPI and IPS300+ collect data from busy intersections while the Campus dataset consists of sparse traffic on campus. Data from different RSU clients are highly diverse according to different sensor devices, sensor deployment, and various scenarios.}
\label{fig:base_dataset}
\vspace{-4mm}
\end{figure*}

\textbf{Base datasets} In intelligent transportation systems, data collected from RSU sensors may differ in various aspects, such as traffic scenes, sensor modality, sensor deployment height and angle, and sensor device models and resolutions. To demonstrate the practical impact and effectiveness of our framework in real-world scenarios, the RSU-SF dataset can greatly reflect the diversity of RSUs. We select three base datasets for RSU perception~\cite{Dair-v2x, busch2022lumpi, wang2022ips300+} and collect another real-world RSU dataset. In total, RSU-SF consists of sensor data collected from 17 RSUs, comprising 21 clients in our dataset. Among these, data of 4 clients originate from vehicle sensors connected to RSUs.

DAIR-V2X~\cite{Dair-v2x} is a large-scale multi-modality dataset for vehicle-to-infrastructure collaborative perception. The dataset comprises point cloud and image data collected from RSUs and vehicles connected to the RSU located at 7 different intersections. Due to the lack of accuracy in the original dataset labeling, we have relabeled all objects' 3D bounding boxes and tracking labels for the infrastructure-side data. 
LUMPI~\cite{busch2022lumpi} is collected using five LiDAR devices deployed at different locations around a single intersection. We consider them as separate RSUs due to their significantly varying perception ranges and resolutions. As the annotations of LUMPI are not yet released, we only use it for training. Specifically, we selected 2,400 frames for training from each RSU. IPS300+~\cite{wang2022ips300+} is a multimodal dataset designed for roadside perception tasks in large-scale urban intersections. The dataset was collected using two intersection perception units equipped with lidar sensors and cameras. Its data is characterized by a high density of objects in the scene, resulting in a high label density and scene complexity. We selected around 3000 frames from the original dataset, out of which 600 frames with complete labels were manually chosen for generating validation and testing pairs.

To increase the scene diversity in our dataset and expand the number of RSUs, we collected additional data from three RSUs on a university campus. 
We selected three distinct locations within the campus setting as the sites for data collection. These sites are situated adjacent to buildings and include the pedestrian pathway or parking lot.
Compared to other traffic scenes, the campus scene has a smaller perception range and a higher proportion of pedestrian and bicycle traffic. 
The LiDAR and camera sensors were deployed at an angle of 45 degrees and a height of 10 meters. More sensor deployment comparisons can be found in Table~\ref{table:sensor_setup}.
We collected about 6,000 frames of multi-modality data, out of which 1,200 samples were annotated for validation and testing.

\begin{table*}[t]
\caption{RSU-SF setup. RSU-SF is constructed from multiple sources, which cover diverse scenarios, devices, angles, heights, and ranges.  *17-20 are four vehicle clients.}
\label{table:sensor_setup}
\vspace{-2mm}
\begin{center}
\begin{tabular}{c|c|cccc|c|c|c}
\toprule
Client Index&0-6&7&8&9&10-11&12-13&14-16 & *17-20\\ 
\midrule
Dataset&DAIR-V2X~\cite{Dair-v2x}& \multicolumn{4}{c|}{LUMPI~\cite{busch2022lumpi}}&IPS300+~\cite{wang2022ips300+}&CAMPUS&DAIR-V2X~\cite{Dair-v2x}\\
Scenario&Crossroads&\multicolumn{4}{c|}{Cusy Intersection}& Busy Intersection&Campus Roadside & Crossroads\\
Modality&Image / PC &PC&PC&PC&PC&PC&Image / PC & PC\\
With GT &w &w/o &w/o &w/o &w/o &w &w & w/o\\
\midrule
\textbf{LiDAR} & & & & & & &  \\
Device&Jaguar&HDL64&Pandar64&PandarQT&VLP16&Ruby-Lite&Leishen C32  &Velodyne 128p\\
Horizontal FOV (\degree)&100&360&360&360&360&360&100 & 100\\
Vertical FOV (\degree)&40&26.5&40&104.2&30&40&31 & 40\\
Channels (beam)&300&64&64&64&16&80&32 &128\\
Rate (Hz)&10&10&10&10&10&10&10 & 10\\
Range (m)&280&120&200&20&100&230&100& 245\\
Range Accuracy (cm)&3&2&2.5&3&3&3&3&3\\
Installation Height (m) &$\sim$6 &$\sim$5 &$\sim$5 &$\sim$5 &$\sim$5 &5.5 & $\sim$6 & {\color{blue} $\sim$2} \\
Installation Angle(\degree) &30 &0 &0 &0 &0 &0 &30&  0\\
\midrule
\textbf{Camera} & & & & & & & \\
Camera Rate(Hz)&25&--&--&--&--&--&20&  --\\
Camera Resolution &1920*1280&--&--&--&--&--&1920*1280&--\\
\bottomrule
\end{tabular}
\vspace{-4mm}
\end{center}
\end{table*}

\subsection{Dataset Properties}

\textbf{ Clients' setup.} In total, the RSU-SF dataset comprises 17 roadside unit clients and 4 vehicle clients, the specific configurations and parameters of each RSU are presented in Table \ref{table:sensor_setup}. It is evident that different RSUs exhibit remarkable diversity in various ways, including scenarios, modalities, LiDAR devices, and installation methods.

The RSU data from DAIR-V2X~\cite{Dair-v2x} is collected from normal traffic crossroads, whereas LUMPI~\cite{busch2022lumpi} and IPS300+~\cite{wang2022ips300+} gather data from busy intersections with exceedingly heavy traffic flow. Differently, our self-collected data is obtained from simpler roadside scenarios on campus.
Furthermore, the utilization of different LiDAR devices and various installation methods gives rise to significant disparities in the resolution and perception range of the point cloud data. we can observe that the perception range of the sensors varies from 20m to 280m.

Additionally, we provide corresponding camera images for certain clients. For RSU units, cameras are cheaper and more easily deployable sensors widely used in traffic surveillance applications. In our proposed framework, we employ image data to generate optical flow, which aids in training the scene flow estimation model.

\begin{figure*}[t]
\centering
\subfloat[Point-Sample]{\includegraphics[width=0.26\textwidth]{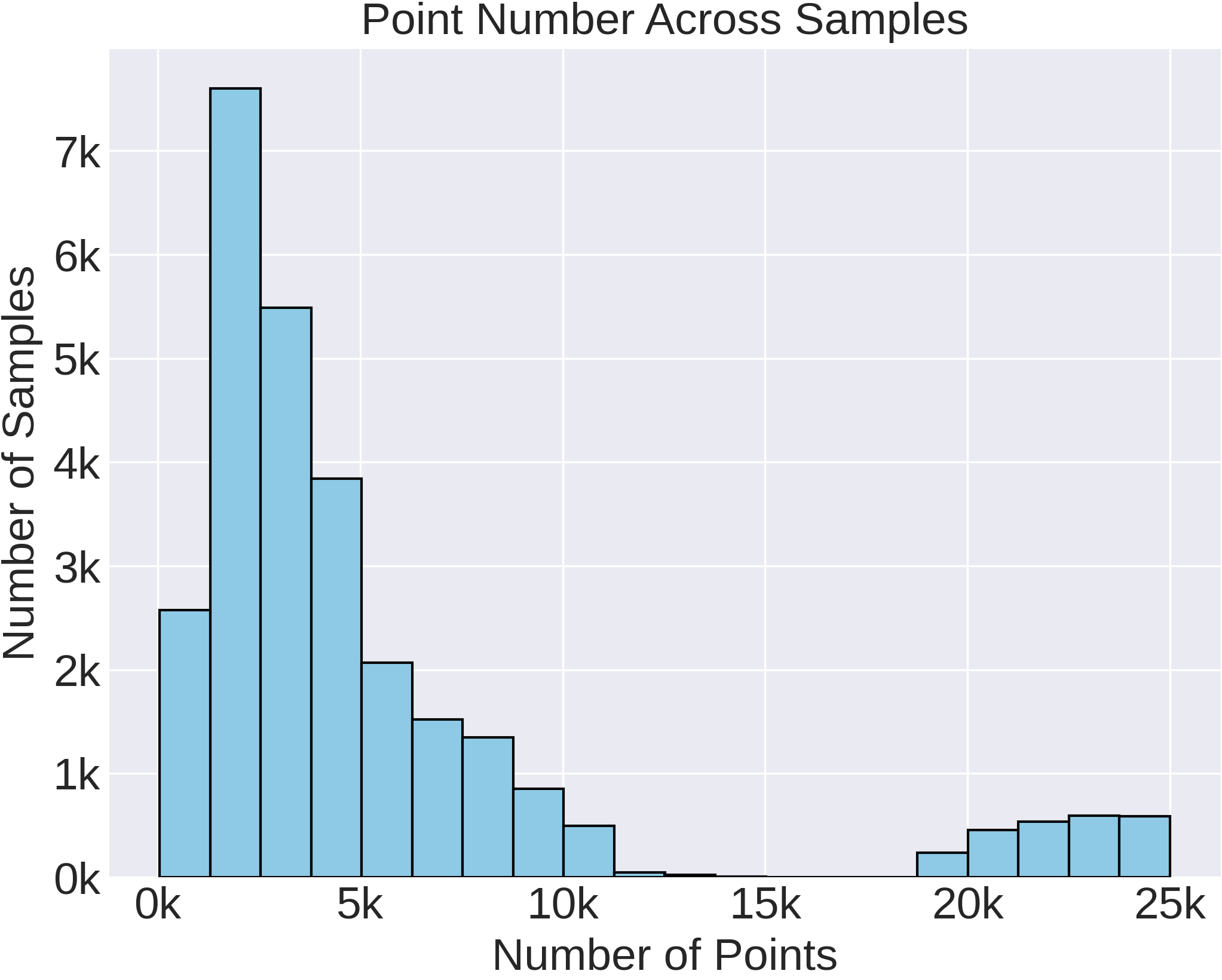}%
\label{fig:data_pc_all}}
\hfil
\subfloat[Flow-Sample]{\includegraphics[width=0.26\textwidth]{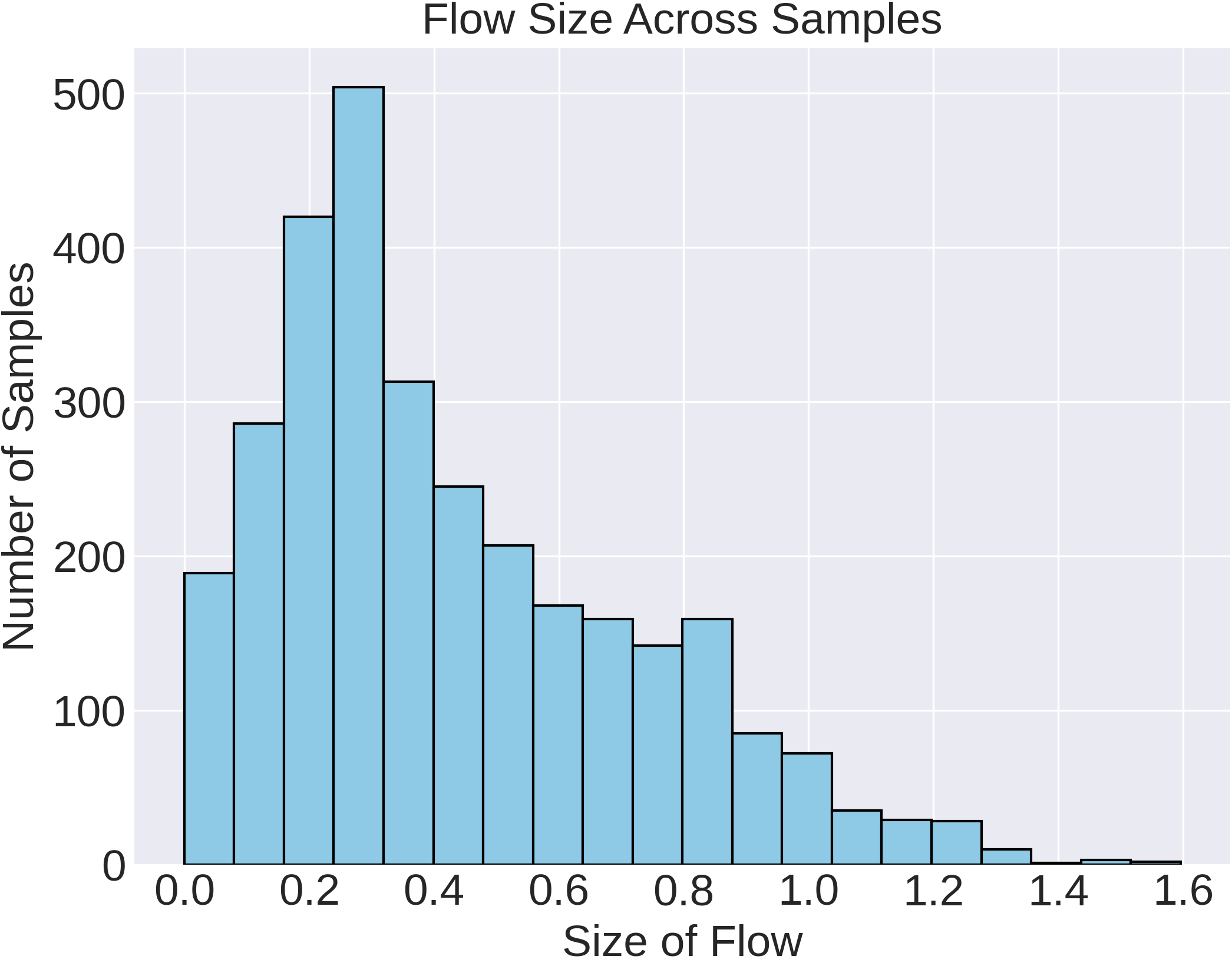}%
\label{fig:data_flow_all}}
\hfil
\subfloat[Dynamic-Sample]{\includegraphics[width=0.26\textwidth]{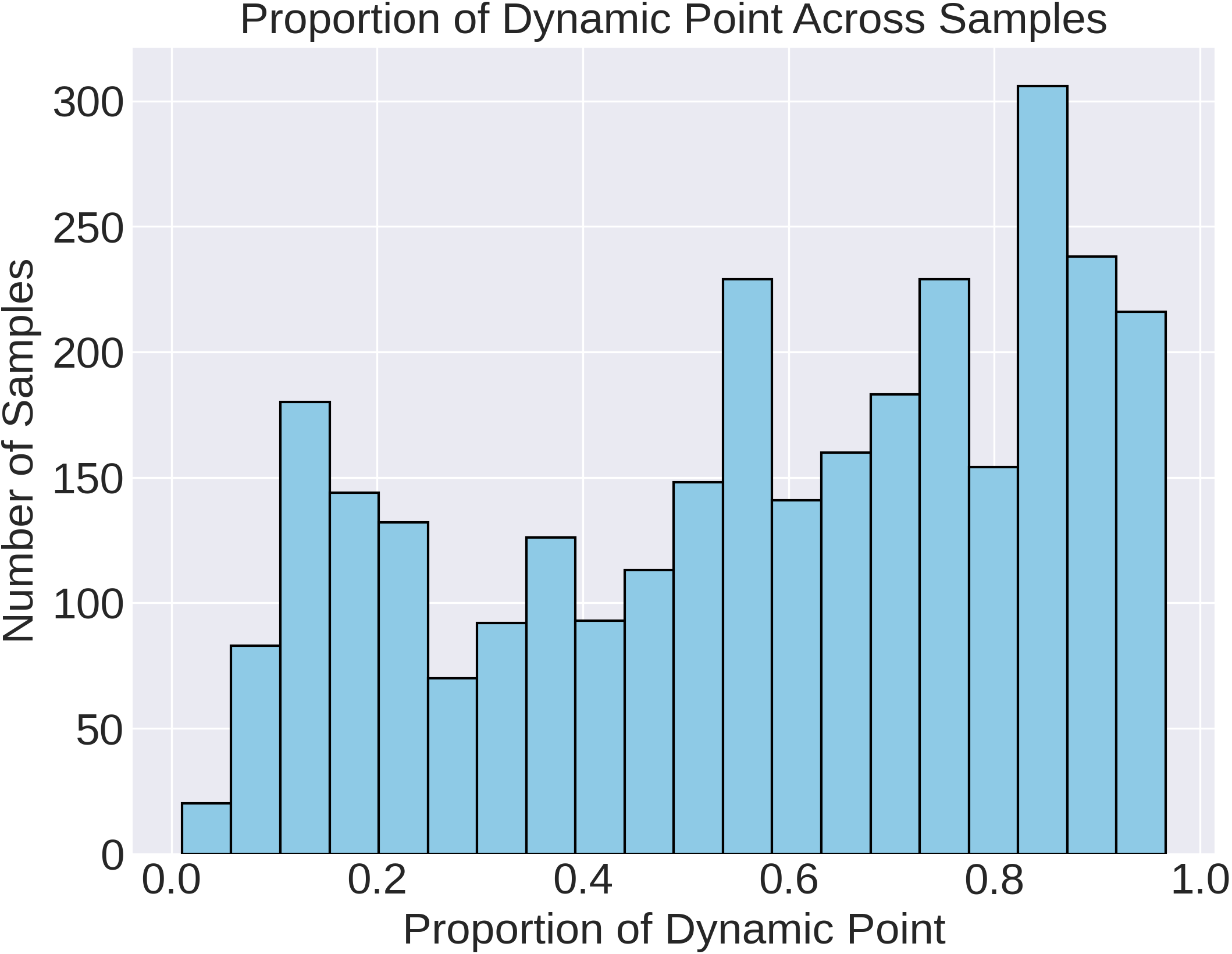}%
\label{fig:data_prop_all}}
\hfil
\subfloat[Point-Client]{\includegraphics[width=0.26\textwidth]{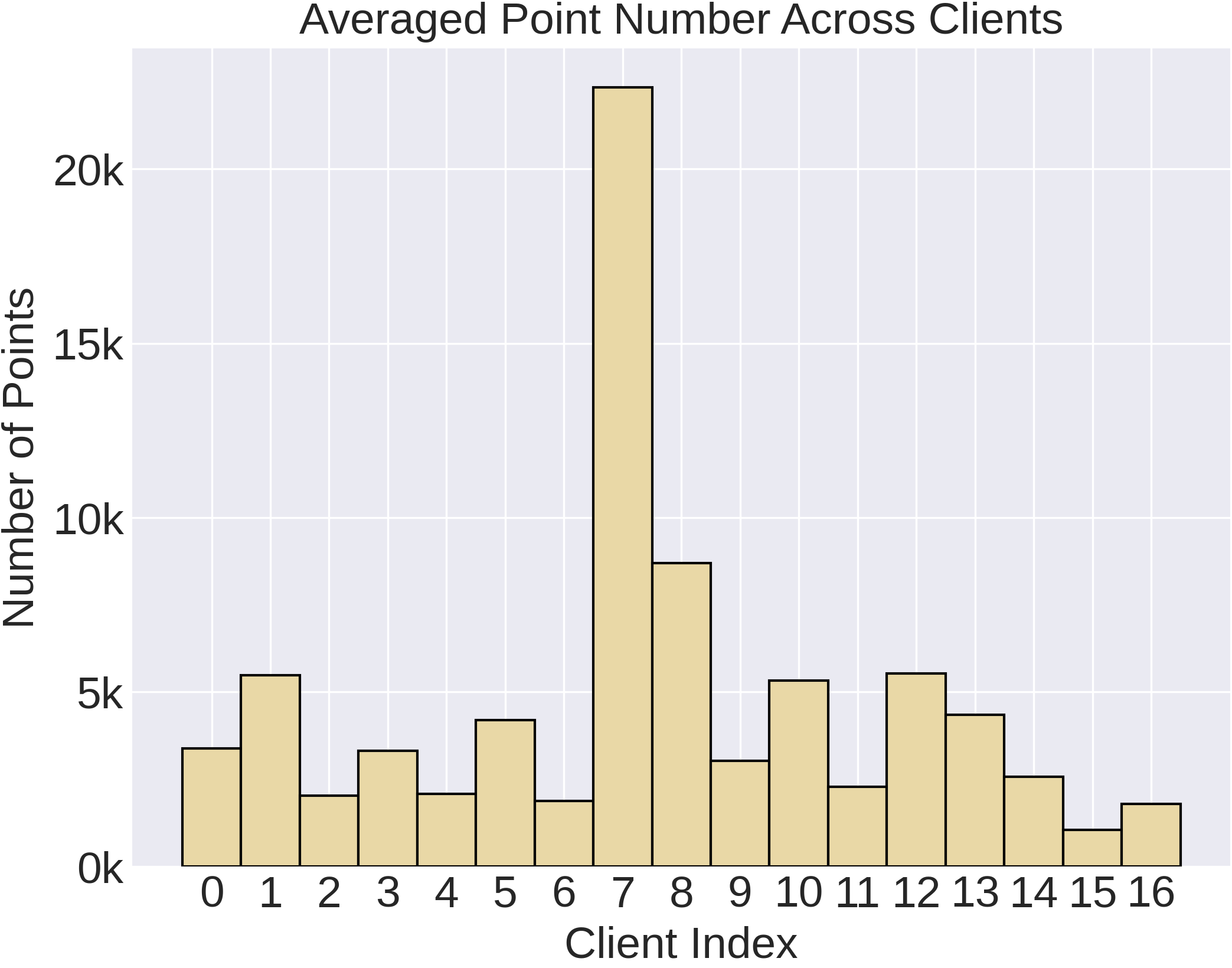}%
\label{fig:data_pc_client}}
\hfil
\subfloat[Flow-Client]{\includegraphics[width=0.26\textwidth]{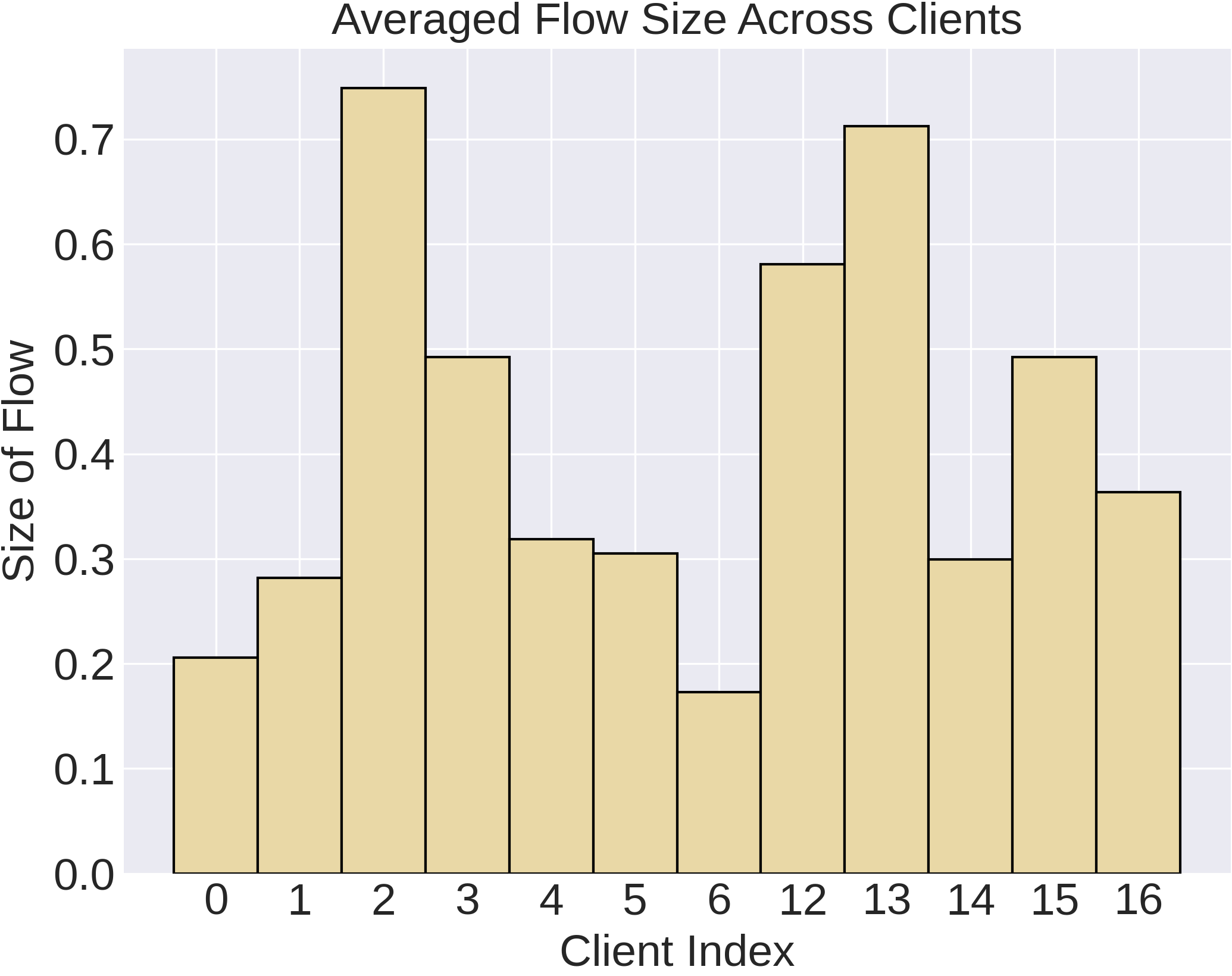}%
\label{fig:data_flow_client}}
\hfil
\subfloat[Dynamic-Client]{\includegraphics[width=0.26\textwidth]{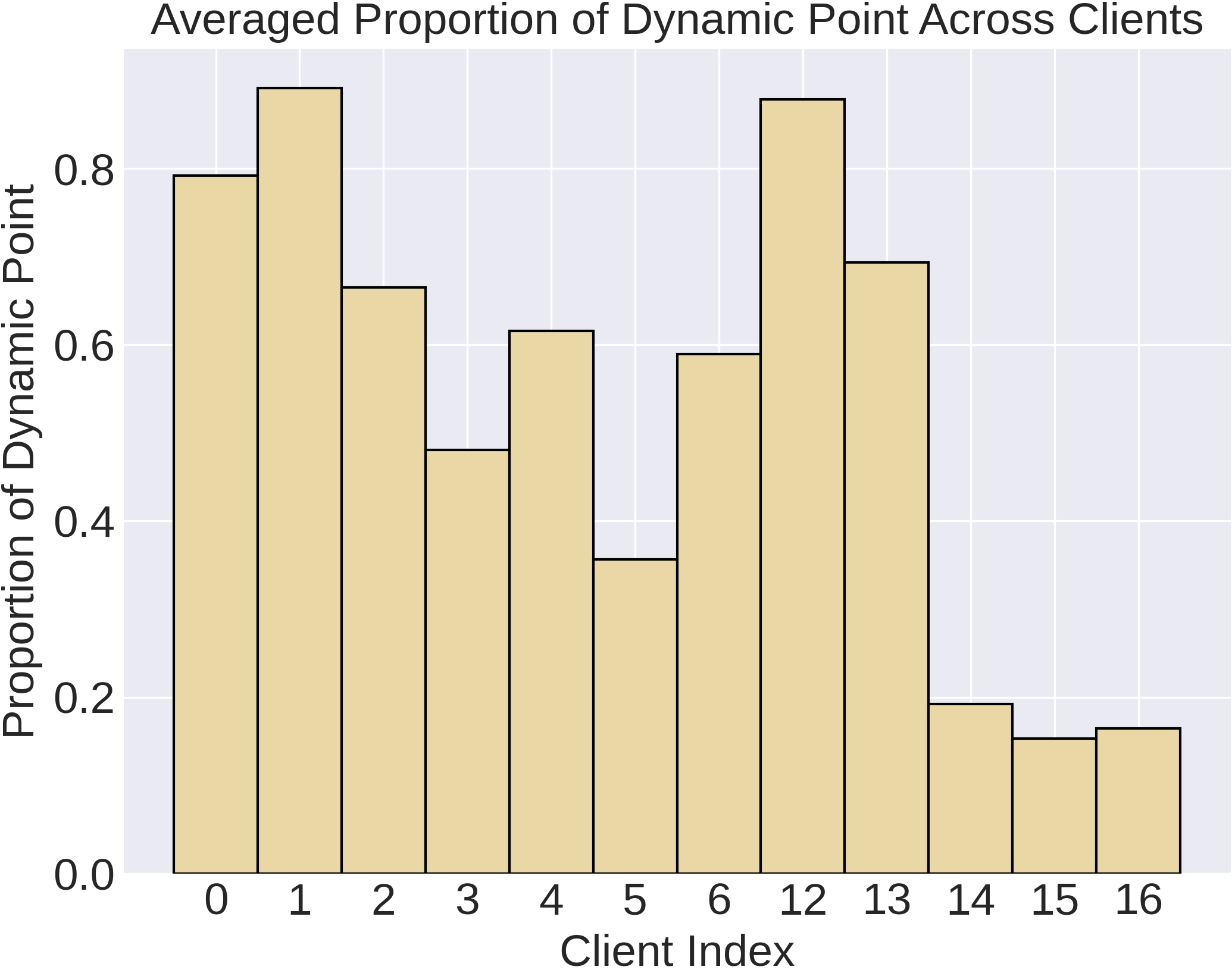}%
\label{fig:data_prop_client}}
\caption{Visualization of the distribution of data properties on two levels (sample-level and client-level). (a) \& (b) show the point number distribution, (c) \& (d) show the flow size distribution, (e) \& (f) show the distribution of dynamic point proportion.}
\label{fig:data}
\vspace{-3mm}
\end{figure*}

\textbf{Statistics.} 
In Fig.~\ref{fig:data}, we provide a quantitative demonstration of sample-level and client-level diversity that exists in the RSU-SF dataset.
Here, we examine from three properties, including the number of point in each data sample, the average size of flow in each data sample, and the proportion of dynamic point in each data sample.
Note that the first is computed on training split and the last two are computed on validation and test splits.
For sample-level demonstration, we plot the histogram where the x-axis is the interested property and the y-axis is the corresponding number of samples (frequency).
For client-level demonstration, we average the property within each client and show the bar plot, where the x-axis is the client index (same as Table~\ref{table:sensor_setup}).

We demonstrate sample-level statistics at the top of Fig.~\ref{fig:data}. 
From the figure, we can see that the samples in RSU-SF dataset cover a wide spectrum for each interested property, which reflects the diversity of RSU-SF.
Specifically, the majority of samples have point counts ranging from $0$ to $10k$, while some samples also exhibit point counts exceeding $20k$.

We demonstrate client-level statistics at the bottom of Fig.~\ref{fig:data}.
Fig.~\ref{fig:data_pc_client} shows the significant difference in the averaged point number of the point cloud data, which is a direct result of the variation in LiDAR devices.
Fig.~\ref{fig:data_flow_client} displays the average ground truth flow value for each client.
It shows that the dynamic patterns (or velocities) of participants vary across clients in different scenarios. 
Moreover, Fig.~\ref{fig:data_prop_client} displays the dynamic point proportions; see display of the category distribution for each client in the Appendix.

Overall, these results show the nature of data heterogeneity of the distributed RSUs' data, emphasizing the need for effective FL algorithm design to train a well-performed model.
Meanwhile, these also indicate the potential of RSU-SF dataset as a new effective benchmark for testing the effectiveness of FL algorithms since there are still limited number of real-world benchmarks for FL.

\begin{table*}[!t]
\caption{Comparisons among representative FL datasets. Five aspects are considered, including the type of partition of clients' data, data modality, task, supervision manner, and the corresponding benchmark.}
\vspace{-2mm}
\label{table:dataset_comp}
\begin{center}
\begin{tabular}{cccccc}
\toprule
Dataset & Partition & Modality & Task & Supervision & FL Benchmark\\
\midrule
Fashion-MNIST~\cite{xiao2017fashion} & Artificial & Image & Classification & Labeled & N/A\\
CIFAR-10/100~\cite{cifar10} & Artificial & Image & Classification & Labeled & N/A\\
Cora/Pubmed/Citeseer~\cite{pflbench} & Artificial & Graph & Classification & Labeled & pFL \\
FEMNIST~\cite{leaf} & Real & Image & Classification & Labeled & N/A\\
Shakespeare~\cite{leaf} & Real & Text & Next-word prediction & Labeled & N/A\\
Cityscapes~\cite{fantauzzo2022feddrive} & Real & Image & Segmentation & Labeled & gFL\\
Fed-LIDC-IDRI~\cite{flamby} & Real & 3D image & Segmentation & Labeled & gFL\\
FLAIR~\cite{flair} & Real & Image & Classification & Labeled & gFL\\
\textbf{RSU-SF (Ours)} & Real & Point cloud \& image & Flow estimation & Unlabeled & gFL \& pFL\\
\bottomrule
\end{tabular}
\vspace{-4mm}
\end{center}
\end{table*}

\textbf{Comparisons with existing datasets.} Table~\ref{table:dataset_comp} shows the comparisons among our constructed RSU-SF and other datasets.
RSU-SF is a practical multi-modal FL dataset with point cloud and image modalities for scene flow estimation task.
Unlike previous works that assume each client has labeled data, our RSU-SF represents more practical scenarios where data is generally unlabeled.
Besides, previous works either emphasize on benchmarking generalized FL or personalized FL, while we simultaneously benchmark generalized and personalized FL.
These properties also make RSU-SF a new and representative test-bed candidate for FL algorithms.
To the best of our knowledge, RSU-SF is the first real-world LiDAR-camera multi-modal dataset and benchmark for the federated learning community.

\section{Experiments}
\label{sec:exp}
\subsection{Implementation Details}

\textbf{Training.} Each experiment is conducted based on PyTorch library on one Nvidia GeForce RTX 3090. We set the number of communication rounds as $50$. Within each round, each client trains the local model for $1$ epoch with a batch size of $32$. The optimizer used is ADAM with a learning rate of $4e^{-4}$. Unless specified, we adopt FlowStep3D~\cite{Flowstep3d} as the local scene flow estimation method. The hyper-parameters of loss terms are $\beta_{ch}=0.75$ and $\beta_{reg}=0.25$, respectively.
For single-modal self-supervised learning, we discard the $s_a$ in loss function~\eqref{eq:cham}.
Except for showing the effectiveness of our multi-modal method, we choose the single-modal version for benchmarking FedRSU to offer a clearer and more foundational perspective on this novel FL framework.
Except for showing the effect of vehicle clients in our method, we use all 17 RSU clients in RSU-SF for experiments.


\textbf{Evaluation.} For more comprehensive comparisons, we consider two evaluation scenarios for general FL methods, namely generalization towards data of seen clients (different data from the same clients as training) and generalization towards data of unseen clients (data from the held-out Campus clients). We consider three metrics, including epe3d, accs, and accr, which are three common scene flow evaluation metrics~\cite{liu2019flownet3d,Flowstep3d,pointpwc}.
(1) epe3d (m): the average end-point-error $||f_{pred} - f_{gt}||_2$ over each point.
(2) accs: percentage of points whose $epe3d < 0.05m$ or relative error $< 5\% $.
(3) accr: percentage of points whose $epe3d < 0.1m$ or relative error $< 10\%$.

\textbf{Self-supervised scene flow methods.} To explore the effects of scene flow methods and provide diverse observations, we consider three representative method candidates (FlowStep3D~\cite{Flowstep3d}, PointPWC-Net~\cite{pointpwc}, FlowNet3D~\cite{liu2019flownet3d}) with the single-modal version of loss function in~\eqref{eq:cham} (discarding $s_a$), concentrating on the different functionality of their backbones. Unless specified, we apply FlowStep3D~\cite{Flowstep3d} as it tends to achieve best in TABLE~\ref{table:self-supervised}.

\begin{table*}[t]
\caption{Generalization evaluation on seen and unseen clients. Two client participation scenarios and three metrics are considered. The lowest epe3d values are \textbf{highlighted} (excluding central learning). Our proposed multi-modal FL approach consistently achieves the lowest epe3d value, which is the most crucial criterion.}
\setlength{\tabcolsep}{2.5pt}
\vspace{-2mm}
\label{tab:mm_generalized}
\begin{center}
\begin{tabular}{l|ccc|ccc|ccc|ccc}
\toprule
Evaluation & \multicolumn{6}{c|}{On Seen Clients} & \multicolumn{6}{c}{On Unseen Clients} \\
Participation & \multicolumn{3}{c|}{Full} & \multicolumn{3}{c|}{Partial} & \multicolumn{3}{c|}{Full} & \multicolumn{3}{c}{Partial}\\
Metric & epe3d ($\downarrow$) & accs ($\uparrow$) & accr ($\uparrow$) & epe3d ($\downarrow$) & accs ($\uparrow$) & accr ($\uparrow$) & epe3d ($\downarrow$) & accs ($\uparrow$) & accr ($\uparrow$) & epe3d ($\downarrow$) & accs ($\uparrow$) & accr ($\uparrow$) \\
\midrule
Central Learning &17.96&30.78&54.70&17.96&30.78&54.70&10.76&37.43&59.32&10.76&37.43&59.32 \\
Local Learning &31.07&30.36&45.67&31.07&30.36&45.67&22.49&25.74&38.50&22.49&25.74&38.50 \\
Local Learning + Multi-Modal
&28.05&32.34&47.23&28.05&32.34&47.23&20.99&28.57&40.32&20.99&28.57&40.32 \\
FedRSU (FedAvg)~\cite{fedavg} &21.69&23.41&47.05&25.28&49.38&60.17&13.75&31.44&52.13&25.57&38.83&42.56 \\
\textbf{FedRSU (FedAvg + M.M.)}&\textbf{20.19}&27.37&49.76&\textbf{20.74}&25.35&47.40&\textbf{9.11}&42.83&63.37&\textbf{9.64}&41.89&62.60 \\
\bottomrule
\end{tabular}
\vspace{-4mm}
\end{center}
\end{table*}

\begin{table*}[!t]
\caption{Personalization evaluation. M.M. denotes multi-modal self-supervision. 1) Mean denotes the metric value mean across clients. 2) Std. denotes the metric value standard deviation across clients. 3) Imp. denotes the ratio of clients whose performance is enhanced through personalized FL (compared with Local). Considering the most crucial criterion (epe3d), our proposed multi-modal FL approach consistently achieves the lowest error (lowest mean), highest fairness (lowest std), and highest incentive (100\% Imp.).}
\vspace{-2mm}
\label{tab:mm_personalized}
\begin{center}
\begin{tabular}{l|ccc|ccc|ccc}
\toprule
\multirow{2}{*}{Evaluation} & \multicolumn{3}{c|}{epe3d} & \multicolumn{3}{c|}{accs} & \multicolumn{3}{c}{accr} \\
& Mean ($\downarrow$) & Std ($\downarrow$) & Imp. ($\uparrow$) & Mean ($\uparrow$) & Std ($\downarrow$) & Imp. ($\uparrow$) & Mean ($\uparrow$) & Std ($\downarrow$) & Imp. ($\uparrow$) \\
\midrule
Central Learning 
&13.93&2.82&85.71&32.45&7.95&71.43&58.47&6.63&100.00 \\
Local Learning 
&21.33&8.34&-&35.83&14.25&-&56.20&15.06&- \\
Local Learing + M.M.
&21.04&6.76&-&20.83&9.07&-&46.42&11.72&-\\
FedRSU (FedAvg + FT)~\cite{fedavg}
&18.49&8.20&71.43&45.94&13.12&100.00&62.86&9.91&100.00 \\
FedRSU (Ditto)~\cite{ditto} 
&19.07&9.23&57.14&39.06&20.59&57.14&58.23&15.12&85.71 \\
\textbf{FedRSU (Ditto + M.M.)}
&15.62&4.36&100.00&30.56&10.56&57.14&58.15&10.61&85.71 \\
\bottomrule
\end{tabular}
\vspace{-4mm}
\end{center}
\end{table*}

\textbf{FL baselines.} We comprehensively consider two lines of FL methods, general FL and personalized FL. 1) For general FL, we consider 7 baselines, including local training, vanilla FL (FedAvg~\cite{fedavg}), methods focusing on local model correction (FedProx~\cite{fedprox}, SCAFFOLD~\cite{scaffold}), and methods focusing on server model adjustment (FedAvgM~\cite{fedavgm}, FedNova~\cite{fednova}, FedAdam~\cite{fedopt}). 2) For personalized FL, we consider 10 baselines, including local training, FedAvg~\cite{fedavg} and FedProx~\cite{fedprox} (with their fine-tuning versions), methods focusing on regularization including Ditto~\cite{ditto}, pFedMe~\cite{pfedme}; methods focusing on partial personalization including FedPer~\cite{fedper} and FedRep~\cite{fedrep}; and pFedGraph~\cite{pfedgraph} that focuses on aggregation.

\subsection{Federated Multi-Modal Learning}

\textbf{Performance on generalized setting.} TABLE~\ref{tab:mm_generalized} compares our proposed federated multi-modal learning approach with classical baselines on the generalized setting, where we consider two classical evaluation protocols and participation scenarios.
From the table, we see that 1) our proposed multi-modal FL method consistently achieves the lowest epe3d across different evaluation protocols and participation scenarios. 
Note that epe3d is the most comprehensive and crucial metric. 
Despite this, our method also achieves the highest or comparable accs and accr except when evaluated on seen clients under partial client participation.
2) Considering the generalization performance (evaluation on unseen clients), our proposed multi-modal FL approach even outperforms the central learning on a single modality, showing the advantages of multi-modality on learning generalizable models.
Note that the unseen clients have fewer dynamic points; see Fig.~\ref{fig:data_prop_client}.
This may indicate that our method can better distinguish static and dynamic points, so that it performs well when switching from training distribution to testing distribution (less dynamic point).

\textbf{Performance on personalized setting.}
TABLE~\ref{tab:mm_personalized} compares our proposed federated multi-modal learning approach with classical baselines on personalized setting, where for each metric we consider three values: mean, standard deviation (std), and improvement ratio (the ratio of clients whose performance is enhanced after joining FL).
Mean value represents the performance, std represents fairness, and improvement ratio captures the level of incentive.
From the table, considering epe3d, our proposed multi-modal FL approach consistently achieves the lowest error (lowest mean), highest fairness (lowest std), and highest level of incentive (highest improvement ratio).

\begin{table*}[t]
\caption{Generalization evaluation on seen and unseen clients. Two client participation scenarios and three metrics are considered.}
\setlength{\tabcolsep}{3pt}
\vspace{-2mm}
\label{table:generalized}
\begin{center}
\begin{tabular}{l|ccc|ccc|ccc|ccc}
\toprule
Evaluation & \multicolumn{6}{c|}{On Seen Clients} & \multicolumn{6}{c}{On Unseen Clients} \\
Participation & \multicolumn{3}{c|}{Full} & \multicolumn{3}{c|}{Partial} & \multicolumn{3}{c|}{Full} & \multicolumn{3}{c}{Partial}\\
Metric & epe3d ($\downarrow$) & accs ($\uparrow$) & accr ($\uparrow$) & epe3d ($\downarrow$) & accs ($\uparrow$) & accr ($\uparrow$) & epe3d ($\downarrow$) & accs ($\uparrow$) & accr ($\uparrow$) & epe3d ($\downarrow$) & accs ($\uparrow$) & accr ($\uparrow$) \\
\midrule
Central Learning &17.96&30.78&54.70&17.96&30.78&54.70&10.76&37.43&59.32&10.76&37.43&59.32 \\
Local Learning &31.07&30.36&45.67&31.07&30.36&45.67&22.49&25.74&38.50&22.49&25.74&38.50 \\
FedRSU (FedAvg)~\cite{fedavg} &21.69&23.41&47.05&25.28&49.38&60.17&13.75&31.44&52.13&25.57&38.83&42.56 \\
FedRSU (FedAvgM)~\cite{fedavgm} &24.18&41.28&56.98&27.83&39.53&55.67&14.34&40.16&51.03&25.42&38.85&42.59 \\
FedRSU (FedProx)~\cite{fedprox} &19.79&31.79&52.34&19.52&21.87&46.78&11.43&42.56&58.00&14.05&41.18&51.48 \\
FedRSU (SCAFFOLD)~\cite{scaffold} &27.29&11.73&54.91&32.13&51.56&59.03&19.42&7.88&46.79&25.51&38.82&42.51 \\
FedRSU (FedNova)~\cite{fednova} &23.66&45.71&59.28&25.66&50.59&59.87&27.09&35.63&42.32&26.15&38.40&42.49 \\
FedRSU (FedAdam)~\cite{fedopt} &28.12&43.27&57.36&29.36&45.48&57.52&30.52&0.84&41.11&26.98&38.26&41.87 \\
\bottomrule
\end{tabular}
\vspace{-4mm}
\end{center}
\end{table*}

\subsection{Benchmark: Generalized Federated Learning}

This section shows experiments on generalized federated learning where the ultimate goal is to train a global model that generalizes well to diverse data. Here, we consider two evaluation protocols on: 1) seen clients and 2) unseen clients. Besides, we experiment on two settings: 1) full client participation, where all clients participate for each round, 2) partial client participation, where only a fraction of clients are available for each round.

\textbf{Generalization towards data of seen clients.} In this experiment, we evaluate the trained global model on the union of held-out test datasets from all the $14$ clients that are involved in the FL training; see results on the left half of Table~\ref{table:generalized}. 

From the table, we see that \textbf{(1)} there is a large gap between local learning and central learning regarding to epe3d, indicating the need of leveraging more data to improve the perception capability of each individual RSU. \textbf{(2)} FedAvg~\cite{fedavg} performs significantly better than local learning, indicating that FL enables improving the perception capability of single RSU through privacy-preserving and communication-efficient collaboration. \textbf{(3)} Generally, partial client participation causes a performance drop, for example, the epe3d value of FedAvg increases from $21.69$ to $25.28$. This indicates that partial client participation could be one practical issue that limits the performance of FedRSU since there could be clients unavailable for each round in practice. \textbf{(4)} The ranking of FL algorithms is surprisingly different from that in existing FL literature on vision tasks, calling for rethinking of FL algorithms designs on different tasks. For example, in vision tasks, SCAFFOLD~\cite{scaffold} tends to perform well~\cite{fednova,feddisco} while FedProx~\cite{fedprox} has rather mediocre performance~\cite{moon,feddisco}. However, we can see that FedProx~\cite{fedprox} performs significantly better than FedAvg~\cite{fedavg} while SCAFFOLD performs worse than FedAvg. Such a finding illustrates that there is no universal FL algorithm for handling all tasks and that existing FL algorithms may not be applicable in FedRSU setting, calling for more specific and effective algorithms in this practical scenario.

\textbf{Generalization towards data of unseen clients.} In this experiment, we evaluate the trained global model on the union of test datasets from the $3$ Campus clients that do not participate the FL training. This experiment is for evaluating the out-of-distribution generalization. From the table, we see that \textbf{(1)} the gap between local learning and central learning becomes larger than that in in-distribution evaluation, indicating that integrating datasets from diverse RSUs can significantly improve the perception capability. Specifically, central learning performs $52\%$ better than local learning. Most FL methods perform better than local training, showing that FL greatly improves over individual learning while preserving privacy. \textbf{(2)} Many methods fail to perform well under partial client participation scenarios, indicating the urgent need for effective methods to handle this common practical scenario.

\begin{table}[t]
\caption{Effects of the number of multi-modal clients (M) in our proposed multi-modal FL approach. Generally, more multi-modal clients indicate better performance.}
\setlength{\tabcolsep}{4pt}
\vspace{-2mm}
\label{table:multimodal}
\begin{center}
\begin{tabular}{l|ccccc}
\toprule
\textbf{Multi-Modal +} & M=7 & M=5 & M=3 & M=1 & M=0 \\
\midrule
Central Learning &13.93&-&-&-&14.59 \\
Local Learning &21.04&21.47&21.11&21.42&21.33 \\
FedRSU (FedAvg)~\cite{fedavg} &15.03&15.40&17.68&18.67&17.28 \\
FedRSU (FedProx)~\cite{fedprox} &14.93&15.04&17.20&17.70&17.84 \\
FedRSU (SCAFFOLD)~\cite{scaffold} &15.14&14.79&14.96&15.19&16.71 \\
\bottomrule
\end{tabular}
\vspace{-4mm}
\end{center}
\end{table}

\textbf{Effects of the number of multi-modal clients in our multi-modal FL approach.}
We have shown the strong performance of our proposed multi-modal FL approach in TABLE~\ref{tab:mm_generalized} and \ref{tab:mm_personalized}.
Further, we explore the effects of the ratio of multi-modal clients to provide deep understanding of the effectiveness of multi-modality.
In TABLE~\ref{table:multimodal}, we randomly discard the image data of some clients, where the experiments are conducted on DAIR-V2X subset of RSU-SF to exclude the effects of other factors.
From the table, we see that the performance generally degrades with the number of multi-modal clients decreases.
This further verifies the effectiveness of our proposed multi-modal training.  The improvement also suggests that camera sensors can substantially reduce the dependency on high-end LiDAR sensors, thereby in practical deployment, cutting costs while maintaining robust detection capabilities.

\subsection{Benchmark: Personalized Federated Learning} The goal of personalized FL is training each personalized model for each client, thus, we evaluate each personalized model on the corresponding held-out test dataset. Here, the experiments are conducted on clients from \cite{Dair-v2x}. For each metric value, we further compare from three angles, including mean value, variance value and improved ratio. The improved ratio denotes the ratio of clients whose performances are enhanced through participating personalization FL. Mean value implies overall utility while variance value and improved ratio reflect the fairness across clients~\cite{lifair}.

From Table~\ref{table:personalized}, we see that \textbf{(1)} the gap between local learning and central learning with regard to mean value of epe3d is large, showing the need to augment individual RSU with outer information. \textbf{(2)} Most FL methods perform better than local learning, indicating that FL enables improving individual utility through collaboration. \textbf{(3)} Personalized FL methods that apply model partition (i.e., FedPer~\cite{fedper} and FedRep~\cite{fedrep}) fail at this setting. This may due to that the patterns in convolutional neural networks (CNNs) for vision domain may not be applicable to our model architecture for lidar domain. In vision domain, there is clear evidence that shallow layers in a CNN capture more general patterns and thus may be more shareable in FL settings. However, this may not hold true in the lidar domain, which may require more explanation for future works and is not the focus of this paper. \textbf{(4)} Considering epe3d, pFedMe~\cite{pfedme} not only achieves the highest utility, but also maintains the highest level of fairness. While pFedMe tends to have mediocre performance in previous literature in vision tasks~\cite{fedala,pfedgraph}. Again, this underlines the need for task-specific algorithms for this setting. Besides, this finding confirms the challenge of designing universal algorithms and calls for more future works to address such challenge.

\begin{table*}[t]
\caption{Personalization evaluation. Mean and std. denote the metric value mean and standard deviation across clients, respectively. Imp. denotes the ratio of clients whose performance is enhanced through personalized FL (compared with Local).}
\vspace{-2mm}
\label{table:personalized}
\begin{center}
\begin{tabular}{l|ccc|ccc|ccc}
\toprule
\multirow{2}{*}{Evaluation} & \multicolumn{3}{c|}{epe3d} & \multicolumn{3}{c|}{accs} & \multicolumn{3}{c}{accr} \\
& Mean ($\downarrow$) & Std. ($\downarrow$) & Imp. ($\uparrow$) & Mean ($\uparrow$) & Std. ($\downarrow$) & Imp. ($\uparrow$) & Mean ($\uparrow$) & Std. ($\downarrow$) & Imp. ($\uparrow$) \\
\midrule
Central Learning &13.93&2.82&85.71&32.45&7.95&71.43&58.47&6.63&100.00 \\
Local Learning &20.09&8.26&-&35.83&14.25&-&56.20&15.06&- \\
FedRSU (FedAvg)~\cite{fedavg} &17.28&8.25&57.14&52.99&11.62&100.00&68.58&7.72&100.00 \\
FedRSU (FedAvg+FT) &18.49&8.20&71.43&45.94&13.12&100.00&62.86&9.91&100.00 \\
FedRSU (FedProx)~\cite{fedprox} &17.84&8.33&57.14&52.06&12.15&100.00&68.30&7.76&100.00 \\
FedRSU (FedProx+FT) &18.00&8.70&57.14&41.97&12.62&85.71&62.32&8.88&71.43 \\
FedRSU (Ditto)~\cite{ditto} &19.07&9.23&57.14&39.06&20.59&57.14&58.23&15.12&85.71 \\
FedRSU (FedPer)~\cite{fedper} &24.07&6.95&14.29&21.08&16.53&0.00&43.06&18.97&0.00 \\
FedRSU (FedRep)~\cite{fedrep} &31.87&7.73&0.00&6.27&8.57&0.00&24.48&18.82&0.00 \\
FedRSU (pFedMe)~\cite{pfedme} &16.38&8.16&85.71&43.63&15.07&85.71&63.50&9.39&100.00 \\
FedRSU (pFedGraph)~\cite{pfedgraph} &17.53&8.25&42.86&46.18&14.59&100.00&64.88&9.32&100.00 \\
\bottomrule
\end{tabular}
\vspace{-4mm}
\end{center}
\end{table*}

\subsection{Effects of Self-supervised Methods}
\label{exp:self}

To explore the effects of self-supervised scene flow estimation methods, we consider three representatives: FlowStep3D~\cite{Flowstep3d}, PointPWC-Net~\cite{pointpwc}, and FlowNet3D~\cite{liu2019flownet3d}. From TABLE~\ref{table:self-supervised}, we see that FlowStep3D outperforms other scene flow methods in terms of epe3d, but it does not necessarily bring better performance on the other two metrics (e.g. FlowNet3D achieves higher accs). Since the metric of epe3d captures the property of all points, we regard it as a more reliable and comprehensive metric, leading to the conclusion that FlowStep3D is a better scene flow method in our FedRSU framework. We believe that with more advanced self-supervised scene flow estimation methods in the future, the effectiveness of our FedRSU framework will be concurrently enhanced.

\begin{table}[t]
\caption{Effects of self-supervised methods in our multi-modal FedRSU method, where FedAvg~\cite{fedavg} is applied. Experiments show that FlowStep3D generally performs the best.}
\vspace{-2mm}
\label{table:self-supervised}
\begin{center}
\begin{tabular}{ccccc}
\toprule
Base Method & Self-Supervision & epe3d & accs & accr \\
\midrule
\multirow{3}{*}{FedRSU + M.M. +} & PointPWC-Net~\cite{pointpwc}& 18.27&55.12&67.36\\
&FlowNet3D~\cite{liu2019flownet3d}& 22.82&56.29&64.00\\
&FlowStep3D~\cite{Flowstep3d}& 17.28&52.99&68.58\\
\bottomrule
\end{tabular}
\vspace{-4mm}
\end{center}
\end{table}

\subsection{Effects of Aggregating Vehicle Sensor Data}
\label{exp:vehicle_side_exp}

Although FedRSU focuses on federated learning on RSUs, technically, with suitable V2X communication technologies, vehicle sensors connected to the RSU could also participate in FedRSU training. Therefore, we explore the effects of incorporating vehicle sensor data into our FedRSU method. The experiments are conducted on the DAIR-V2X subset of the RSU-SF dataset to control for external factors. In Table~\ref{table:vehicle_client}, we gradually increase the number of vehicle clients involved in the FedRSU training (from V=0 to V=4) by randomly introducing vehicle client data from the RSU-SF dataset. Experimental results show that directly incorporating vehicle client data into the FedRSU method's training negatively impacts the overall model performance. This is due to significant differences in data characteristics between RSU and vehicle sensors: First, the deployment of RSU and vehicle sensors differ substantially, as shown in Table~\ref{table:sensor_setup}, including differences in height and orientation. Second, RSU sensors are statically deployed, while vehicles are mobile, leading to significant noise in the point cloud data collected by vehicle LiDAR sensors. Addressing these challenges and effectively integrating vehicle data into FedRSU training is a challenging problem, and we leave it to future research.

\begin{table}[t]
\caption{Effects of the number of vehicle clients (V) in our FedRSU method, where FedAvg~\cite{fedavg} is applied.}
\setlength{\tabcolsep}{4pt}
\vspace{-2mm}
\label{table:vehicle_client}
\begin{center}
\begin{tabular}{l|ccccc}
\toprule
\textbf{Vehicle clients} & V=0 & V=1 & V=2 & V=3 & V=4 \\
\midrule
Central Learning &14.59&-&-&-& 18.09 \\
FedRSU (FedAvg)~\cite{fedavg} &\textbf{17.28}& 18.75 & 20.69 & 20.09 & 19.80 \\

\bottomrule
\end{tabular}
\vspace{-4mm}
\end{center}
\end{table}

\subsection{Visualization of Predicted Flow}

\begin{figure*}[t]
\centering
\subfloat[Central Learning]{\includegraphics[width=0.3\textwidth]{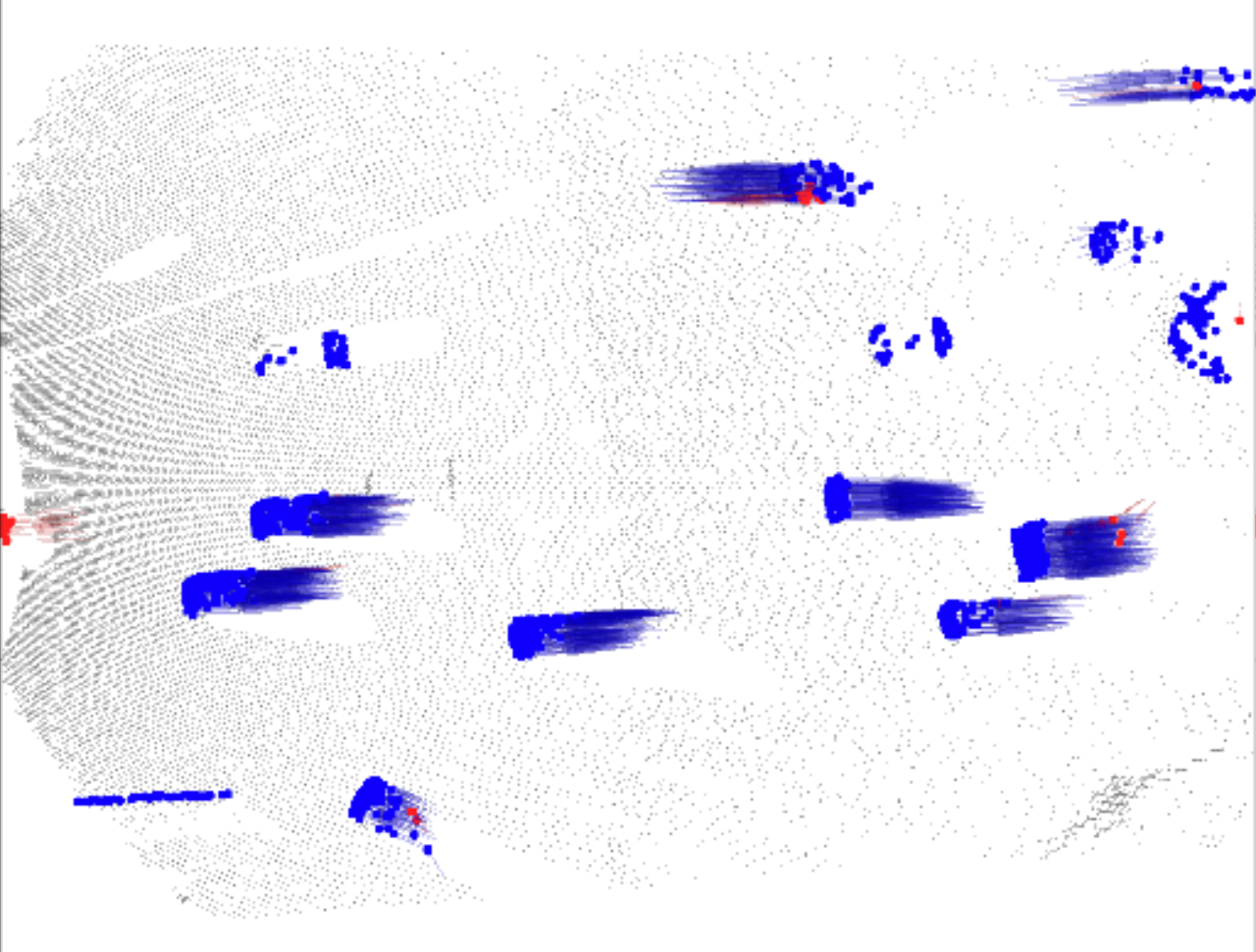}%
\label{fig:vis_central}}
\hfil
\subfloat[Federated Learning]{\includegraphics[width=0.3\textwidth]{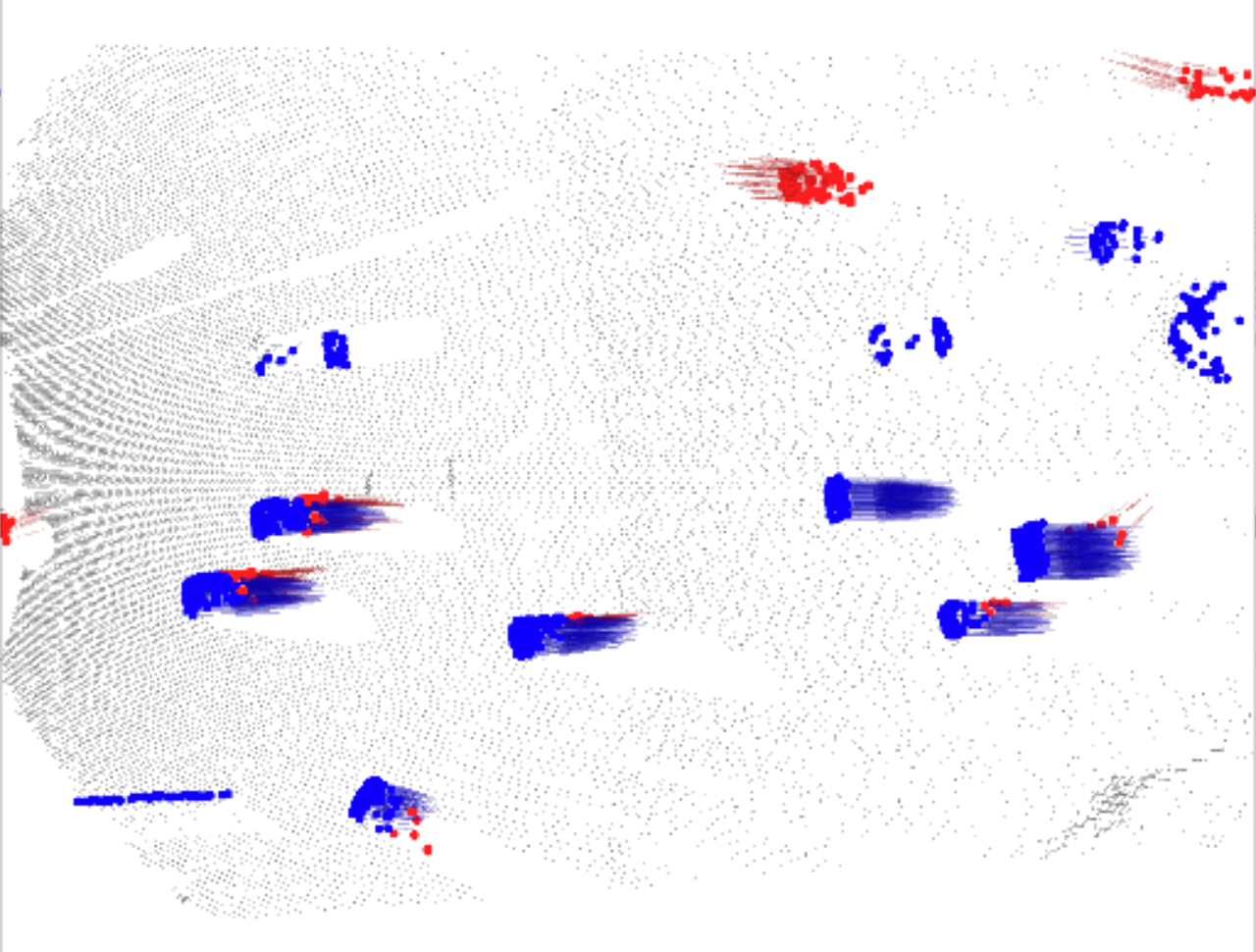}%
\label{fig:vis_fed}}
\hfil
\subfloat[Local Learning]{\includegraphics[width=0.3\textwidth]{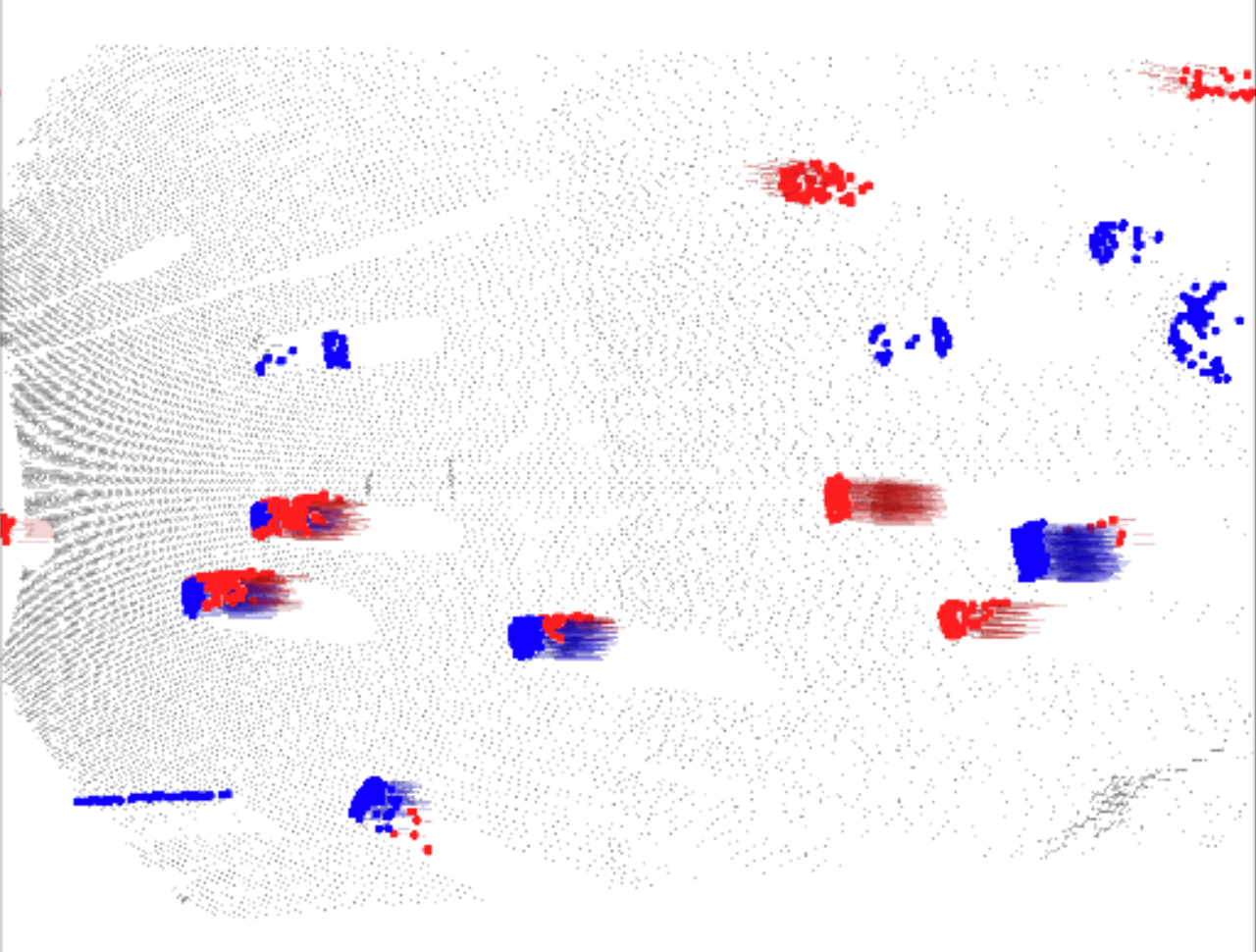}%
\label{fig:vis_local}}
\caption{Qualitative analysis via visualization of source and predicted point cloud. Black point denotes source point cloud, blue point denotes correct point ($epe3d<0.3$) in the predicted cloud, and red point ($epe3d \geq 0.3$) denotes false point in the predicted cloud. We can see that federated learning significantly outperforms local learning while there is still a gap between central learning and federated learning.}
\label{fig:visualization}
\vspace{-4mm}
\end{figure*}

For deeper understanding of the task, we visualize the prediction capability for qualitative analysis. Following the setting full client participation setting in Table~\ref{table:generalized}, we evaluate the predicted results of three representative methods, including central learning, federated learning (FedAvg~\cite{fedavg}), and local learning. Here, we show the results on a sample in Fig.~\ref{fig:visualization}, where black dots denote source point cloud, blue dots denote well-predicted points and red dots denote poorly-predicted points.

From the figure, we see that 1) local (individual) learning may result in poor scene flow estimation due to limited perceived data from a single RSU; see a large proportion of red dots on the right of Fig.~\ref{fig:visualization}. 2) Federated learning significantly improves the quality of scene flow estimation; see a large proportion of poorly-predicted red points in local learning have transformed into well-predicted blue points in federated learning (e.g., the two clusters in left of Fig.~\ref{fig:vis_fed}). 3) There is still a gap between federated learning and central learning; see better prediction of central learning in the middle top of Fig.~\ref{fig:vis_central}. These results illustrate that FL can improve the prediction ability over non-collaborating individual learning. However, data heterogeneity still adversely hinders FL from performing as effectively as central learning, thus calling for more effective algorithm designs for FedRSU scenarios.

\section{Discussion}
\label{sec:discuss}

\subsection{Real-world Deployment in Large Scale}

One concern may be the high cost of LiDAR sensors when deployed on a large scale. However, considering the following points, this is not a severe issue. 
First, using LiDAR sensors in the RSU setting is prevalent, including many datasets and algorithms developed based on these datasets, as discussed in Section~\ref{sec:rsu_related_work}. The usage and promotion of RSU LiDARs in practical applications is a growing trend. 
Second, from a cost perspective, the price of long-range LiDAR sensors has been consistently decreasing. Commercially, RSU perception systems with LiDAR sensors are already available from companies like Leishen and Robosense. 
Third, our proposed multi-modality approach can reduce dependency on expensive LiDAR sensors. By including camera data, our method can enhance overall performance by 15\%, which suggests that camera sensors can substantially reduce the dependency on high-end LiDAR sensors, thereby cutting costs while maintaining robust detection capabilities.

\subsection{Research Directions}

\textbf{Data heterogeneity.} One essential future direction is tackling LiDAR data heterogeneity to achieve better scene flow prediction. Since many previous FL works are explored under the context of artificially-construction image classification scenarios~\cite{fedrod}, their performance and applicability are not guaranteed in the proposed real-world FedRSU scenario, calling for more practical and scenario-oriented algorithms.

\textbf{Multi-modality and modality heterogeneity.} FedRSU opens an interesting yet under-explored multi-modality scenario for FL. Most previous FL works focus on uni-modality tasks, including image classification~\cite{flair}, text classification~\cite{leaf}, and image segmentation~\cite{flamby}, while multi-modality is also common in real world and has gained lots of attention in the centralized scope~\cite{clip,gpt4}. Though we regard camera data as additional supervision in this paper, there are still many potential research directions, including multi-modality fusion and mitigating the issues of missing modality.

Additionally, while the FedRSU system primarily considers RSU sensor data, vehicle sensors connected to RSUs can also be a substantial source of training data. Despite our dataset providing both RSU and vehicle data, incorporating vehicle data directly into FedRSU training does not effectively improve the model’s performance under the current method design. This lack of efficacy is mainly due to the differences in data characteristics between RSU and vehicle sensors. Bridging this gap and effectively combining these two types of data calls for additional methodological design.

\textbf{Availability heterogeneity.} FedRSU is also faced with the issue of availability heterogeneity, where RSUs from different regions may have different connecting stability (e.g., RSUs from urban and rural areas). This issue could be critical since the final global model may fail to perform well on RSUs in rural areas if most of the participating RSUs are from urban areas. To ensure balanced improvement across different areas of the transportation system, this issue should be fairly considered and alleviated.

\textbf{System heterogeneity.} FedRSU may encounter the issue of system heterogeneity, where the computing speeds of different RSUs could be distinct. This issue may result in different numbers of local model updates at each round, leading to objective inconsistency problem in FL~\cite{fednova}. Previous works often focus on imbalanced image classification tasks, whose behaviors are unforeseeable for this scene flow estimation task on point cloud data.

\textbf{Model heterogeneity.} Model heterogeneity could be a natural and interesting research direction in FedRSU, where different clients may conduct training on different model sizes and architectures due to the different computing capabilities and memory spaces of different RSUs. This issue makes the conventional model aggregation techniques inapplicable, calling for new techniques (e.g., knowledge distillation and model pruning) for sharing knowledge among clients under the context of self-supervised tasks.

\subsection{Limitations}

\textbf{Limited client number.} Though we have made great efforts on collecting data and annotating the scene (for evaluation), the total client number is still somewhat limited due to the high cost of time and resources. Except that we will continuously expand this dataset, one possible solution is that one can split the current dataset for simulation by techniques such as rotating with different degrees or sampling with different ratios.

\textbf{Scene flow estimation task.}
Since in the intelligent transportation system, it is challenging to obtain ground truth annotations at the local (RSU) end, we choose self-supervised scene flow estimation as our target task. The scene flow estimation doesn’t offer a holistic scene representation. It is a low-level task that anticipates the dynamic flow within the scene, making it intricate to distinguish object categories and stationary instances. For practical usage, it typically requires integration with other downstream tasks. 

Nevertheless, our proposed system boasts the flexibility to utilize disparate tasks and learning frameworks. With the current advancements in point-cloud pretraining~\cite{li2022simipu, min2022voxel} and auto-labeling technologies~\cite{najibi2022motion, zhang2023towards}, we now have more choices for practical task learning that are independent of manual labels. In our framework, model training at the local end can be easily replaced with these models and methodologies. Moreover, our system has revealed the potential of expanding these model training paradigms to significantly larger scalability in an easy and economical way.

\section{Conclusions}
\label{sec:conclude}

In this paper, we propose a new practical scenario for intelligent transportation systems (ITS): FedRSU, where multiple roadside units (RSUs) collaboratively train a shared scene flow estimation model using the continuously generated unlabeled point cloud (and camera) data.
In FedRSU, we propose a novel multi-modal FL approach that trains the scene flow estimation model on point cloud under fine-grained guidance from image optical flow.
To verify the performance of FedRSU, we construct a real-world scene flow dataset RSU-SF, which covers diverse scenarios, devices and configurations, promoting the development of FedRSU and FL.
Additionally, we provide a comprehensive benchmark on FedRSU, covering diverse baselines and scenarios, and show the outstanding performance of our proposed multi-modal FL approach.

We show that FedRSU can remedy the limitation of a single RSU, augmenting ITS with more accurate, wide-range, and reliable perception capabilities.
Our work demonstrates an easy and deployable framework to scale up training datasets to a large extent for model learning in ITS.
Besides, our work presents a new real-world scenario and opens new research directions for the FL community.

\section*{Acknowledgements}
This research is supported by the National Key R\&D Program of China under Grant 2021ZD0112801, NSFC under Grant 62171276 and the Science and Technology Commission of Shanghai Municipal under Grant 21511100900 and 22DZ2229005.

\bibliographystyle{IEEEtran}
\bibliography{ref}

\begin{thebibliography}{100}
\providecommand{\url}[1]{#1}
\csname url@samestyle\endcsname
\providecommand{\newblock}{\relax}
\providecommand{\bibinfo}[2]{#2}
\providecommand{\BIBentrySTDinterwordspacing}{\spaceskip=0pt\relax}
\providecommand{\BIBentryALTinterwordstretchfactor}{4}
\providecommand{\BIBentryALTinterwordspacing}{\spaceskip=\fontdimen2\font plus
\BIBentryALTinterwordstretchfactor\fontdimen3\font minus \fontdimen4\font\relax}
\providecommand{\BIBforeignlanguage}[2]{{%
\expandafter\ifx\csname l@#1\endcsname\relax
\typeout{** WARNING: IEEEtran.bst: No hyphenation pattern has been}%
\typeout{** loaded for the language `#1'. Using the pattern for}%
\typeout{** the default language instead.}%
\else
\language=\csname l@#1\endcsname
\fi
#2}}
\providecommand{\BIBdecl}{\relax}
\BIBdecl

\bibitem{guerna2022roadside}
A.~Guerna, S.~Bitam, and C.~T. Calafate, ``Roadside unit deployment in internet of vehicles systems: A survey,'' \emph{Sensors}, vol.~22, no.~9, p. 3190, 2022.

\bibitem{magsino2022enhanced}
E.~R. Magsino and I.~W.-H. Ho, ``An enhanced information sharing roadside unit allocation scheme for vehicular networks,'' \emph{IEEE Transactions on Intelligent Transportation Systems}, vol.~23, no.~9, pp. 15\,462--15\,475, 2022.

\bibitem{ackels2021survey}
S.~Ackels, P.~Benavidez, and M.~Jamshidi, ``A survey of modern roadside unit deployment research,'' in \emph{2021 World Automation Congress (WAC)}.\hskip 1em plus 0.5em minus 0.4em\relax IEEE, 2021, pp. 7--14.

\bibitem{chen2017vehicle}
S.~Chen, J.~Hu, Y.~Shi, Y.~Peng, J.~Fang, R.~Zhao, and L.~Zhao, ``Vehicle-to-everything (v2x) services supported by lte-based systems and 5g,'' \emph{IEEE Communications Standards Magazine}, vol.~1, no.~2, pp. 70--76, 2017.

\bibitem{xu2022v2x}
R.~Xu, H.~Xiang, Z.~Tu, X.~Xia, M.-H. Yang, and J.~Ma, ``V2x-vit: Vehicle-to-everything cooperative perception with vision transformer,'' in \emph{European conference on computer vision}.\hskip 1em plus 0.5em minus 0.4em\relax Springer, 2022, pp. 107--124.

\bibitem{hu2022where2comm}
Y.~Hu, S.~Fang, Z.~Lei, Y.~Zhong, and S.~Chen, ``Where2comm: Communication-efficient collaborative perception via spatial confidence maps,'' \emph{Advances in neural information processing systems}, vol.~35, pp. 4874--4886, 2022.

\bibitem{ren2023interruption}
S.~Ren, Z.~Lei, Z.~Wang, M.~Dianati, Y.~Wang, S.~Chen, and W.~Zhang, ``Interruption-aware cooperative perception for v2x communication-aided autonomous driving,'' \emph{arXiv preprint arXiv:2304.11821}, 2023.

\bibitem{Dair-v2x}
H.~Yu, Y.~Luo, M.~Shu, Y.~Huo, Z.~Yang, Y.~Shi, Z.~Guo, H.~Li, X.~Hu, J.~Yuan \emph{et~al.}, ``Dair-v2x: A large-scale dataset for vehicle-infrastructure cooperative 3d object detection,'' in \emph{Proceedings of the IEEE/CVF Conference on Computer Vision and Pattern Recognition}, 2022, pp. 21\,361--21\,370.

\bibitem{busch2022lumpi}
S.~Busch, C.~Koetsier, J.~Axmann, and C.~Brenner, ``Lumpi: The leibniz university multi-perspective intersection dataset,'' in \emph{2022 IEEE Intelligent Vehicles Symposium (IV)}.\hskip 1em plus 0.5em minus 0.4em\relax IEEE, 2022, pp. 1127--1134.

\bibitem{wang2022ips300+}
H.~Wang, X.~Zhang, Z.~Li, J.~Li, K.~Wang, Z.~Lei, and R.~Haibing, ``Ips300+: a challenging multi-modal data sets for intersection perception system,'' in \emph{2022 International Conference on Robotics and Automation (ICRA)}.\hskip 1em plus 0.5em minus 0.4em\relax IEEE, 2022, pp. 2539--2545.

\bibitem{Rope3d}
X.~Ye, M.~Shu, H.~Li, Y.~Shi, Y.~Li, G.~Wang, X.~Tan, and E.~Ding, ``Rope3d: The roadside perception dataset for autonomous driving and monocular 3d object detection task,'' in \emph{Proceedings of the IEEE/CVF Conference on Computer Vision and Pattern Recognition}, 2022, pp. 21\,341--21\,350.

\bibitem{A9-I}
W.~Zimmer, C.~Cre{\ss}, H.~T. Nguyen, and A.~C. Knoll, ``A9 intersection dataset: All you need for urban 3d camera-lidar roadside perception,'' \emph{arXiv preprint arXiv:2306.09266}, 2023.

\bibitem{he2022masked}
K.~He, X.~Chen, S.~Xie, Y.~Li, P.~Doll{\'a}r, and R.~Girshick, ``Masked autoencoders are scalable vision learners,'' in \emph{Proceedings of the IEEE/CVF conference on computer vision and pattern recognition}, 2022, pp. 16\,000--16\,009.

\bibitem{devlin2018bert}
J.~Devlin, M.-W. Chang, K.~Lee, and K.~Toutanova, ``Bert: Pre-training of deep bidirectional transformers for language understanding,'' \emph{arXiv preprint arXiv:1810.04805}, 2018.

\bibitem{vedula1999three}
S.~Vedula, S.~Baker, P.~Rander, R.~Collins, and T.~Kanade, ``Three-dimensional scene flow,'' in \emph{Proceedings of the Seventh IEEE International Conference on Computer Vision}, vol.~2.\hskip 1em plus 0.5em minus 0.4em\relax IEEE, 1999, pp. 722--729.

\bibitem{fortun2015optical}
D.~Fortun, P.~Bouthemy, and C.~Kervrann, ``Optical flow modeling and computation: A survey,'' \emph{Computer Vision and Image Understanding}, vol. 134, pp. 1--21, 2015.

\bibitem{slim}
S.~A. Baur, D.~J. Emmerichs, F.~Moosmann, P.~Pinggera, B.~Ommer, and A.~Geiger, ``Slim: Self-supervised lidar scene flow and motion segmentation,'' in \emph{Proceedings of the IEEE/CVF International Conference on Computer Vision}, 2021, pp. 13\,126--13\,136.

\bibitem{huang2022representation}
X.~Huang, Y.~Wang, V.~Guizilini, R.~Ambrus, A.~Gaidon, and J.~Solomon, ``Representation learning for object detection from unlabeled point cloud sequences,'' in \emph{Conference on Robot Learning}, 2022.

\bibitem{pillar_motion}
C.~Luo, X.~Yang, and A.~Yuille, ``Self-supervised pillar motion learning for autonomous driving,'' in \emph{Proceedings of the IEEE/CVF Conference on Computer Vision and Pattern Recognition}, 2021, pp. 3183--3192.

\bibitem{fang2024self}
S.~Fang, Z.~Liu, M.~Wang, C.~Xu, Y.~Zhong, and S.~Chen, ``Self-supervised bird's eye view motion prediction with cross-modality signals,'' \emph{arXiv preprint arXiv:2401.11499}, 2024.

\bibitem{mittal2020just}
H.~Mittal, B.~Okorn, and D.~Held, ``Just go with the flow: Self-supervised scene flow estimation,'' in \emph{Proceedings of the IEEE/CVF conference on computer vision and pattern recognition}, 2020, pp. 11\,177--11\,185.

\bibitem{pointpwc}
W.~Wu, Z.~Y. Wang, Z.~Li, W.~Liu, and L.~Fuxin, ``Pointpwc-net: Cost volume on point clouds for (self-) supervised scene flow estimation,'' in \emph{Computer Vision--ECCV 2020: 16th European Conference, Glasgow, UK, August 23--28, 2020, Proceedings, Part V 16}.\hskip 1em plus 0.5em minus 0.4em\relax Springer, 2020, pp. 88--107.

\bibitem{Flowstep3d}
Y.~Kittenplon, Y.~C. Eldar, and D.~Raviv, ``Flowstep3d: Model unrolling for self-supervised scene flow estimation,'' in \emph{Proceedings of the IEEE/CVF Conference on Computer Vision and Pattern Recognition}, 2021, pp. 4114--4123.

\bibitem{weng2022s2net}
X.~Weng, J.~Nan, K.-H. Lee, R.~McAllister, A.~Gaidon, N.~Rhinehart, and K.~M. Kitani, ``S2net: Stochastic sequential pointcloud forecasting,'' in \emph{European Conference on Computer Vision}.\hskip 1em plus 0.5em minus 0.4em\relax Springer, 2022, pp. 549--564.

\bibitem{khurana2023point}
T.~Khurana, P.~Hu, D.~Held, and D.~Ramanan, ``Point cloud forecasting as a proxy for 4d occupancy forecasting,'' in \emph{Proceedings of the IEEE/CVF Conference on Computer Vision and Pattern Recognition}, 2023, pp. 1116--1124.

\bibitem{advances}
P.~Kairouz, H.~B. McMahan, B.~Avent, A.~Bellet, M.~Bennis, A.~N. Bhagoji, K.~Bonawitz, Z.~Charles, G.~Cormode, R.~Cummings \emph{et~al.}, ``Advances and open problems in federated learning,'' \emph{arXiv preprint arXiv:1912.04977}, 2019.

\bibitem{yang_survey}
Q.~Yang, Y.~Liu, T.~Chen, and Y.~Tong, ``Federated machine learning: Concept and applications,'' \emph{ACM Transactions on Intelligent Systems and Technology (TIST)}, vol.~10, no.~2, pp. 1--19, 2019.

\bibitem{li_survey}
T.~Li, A.~K. Sahu, A.~Talwalkar, and V.~Smith, ``Federated learning: Challenges, methods, and future directions,'' \emph{IEEE Signal Processing Magazine}, vol.~37, no.~3, pp. 50--60, 2020.

\bibitem{rjoub2021improving}
G.~Rjoub, O.~A. Wahab, J.~Bentahar, and A.~S. Bataineh, ``Improving autonomous vehicles safety in snow weather using federated yolo cnn learning,'' in \emph{Mobile Web and Intelligent Information Systems: 17th International Conference, MobiWIS 2021, Virtual Event, August 23--25, 2021, Proceedings}.\hskip 1em plus 0.5em minus 0.4em\relax Springer, 2021, pp. 121--134.

\bibitem{wang2022federated}
S.~Wang, C.~Li, D.~W.~K. Ng, Y.~C. Eldar, H.~V. Poor, Q.~Hao, and C.~Xu, ``Federated deep learning meets autonomous vehicle perception: Design and verification,'' \emph{IEEE network}, 2022.

\bibitem{liu2020fedvision}
Y.~Liu, A.~Huang, Y.~Luo, H.~Huang, Y.~Liu, Y.~Chen, L.~Feng, T.~Chen, H.~Yu, and Q.~Yang, ``Fedvision: An online visual object detection platform powered by federated learning,'' in \emph{Proceedings of the AAAI Conference on Artificial Intelligence}, vol.~34, no.~08, 2020, pp. 13\,172--13\,179.

\bibitem{fantauzzo2022feddrive}
L.~Fantauzzo, E.~Fan{\`\i}, D.~Caldarola, A.~Tavera, F.~Cermelli, M.~Ciccone, and B.~Caputo, ``Feddrive: generalizing federated learning to semantic segmentation in autonomous driving,'' in \emph{2022 IEEE/RSJ International Conference on Intelligent Robots and Systems (IROS)}.\hskip 1em plus 0.5em minus 0.4em\relax IEEE, 2022, pp. 11\,504--11\,511.

\bibitem{shenaj2023learning}
D.~Shenaj, E.~Fan{\`\i}, M.~Toldo, D.~Caldarola, A.~Tavera, U.~Michieli, M.~Ciccone, P.~Zanuttigh, and B.~Caputo, ``Learning across domains and devices: Style-driven source-free domain adaptation in clustered federated learning,'' in \emph{Proceedings of the IEEE/CVF Winter Conference on Applications of Computer Vision}, 2023, pp. 444--454.

\bibitem{song2023fedbevt}
R.~Song, R.~Xu, A.~Festag, J.~Ma, and A.~Knoll, ``Fedbevt: Federated learning bird's eye view perception transformer in road traffic systems,'' \emph{arXiv preprint arXiv:2304.01534}, 2023.

\bibitem{zhang2021real}
H.~Zhang, J.~Bosch, and H.~H. Olsson, ``Real-time end-to-end federated learning: An automotive case study,'' in \emph{2021 IEEE 45th Annual Computers, Software, and Applications Conference (COMPSAC)}.\hskip 1em plus 0.5em minus 0.4em\relax IEEE, 2021, pp. 459--468.

\bibitem{zhang2021end}
------, ``End-to-end federated learning for autonomous driving vehicles,'' in \emph{2021 International Joint Conference on Neural Networks (IJCNN)}.\hskip 1em plus 0.5em minus 0.4em\relax IEEE, 2021, pp. 1--8.

\bibitem{han2022federated}
M.~Han, K.~Xu, S.~Ma, A.~Li, and H.~Jiang, ``Federated learning-based trajectory prediction model with privacy preserving for intelligent vehicle,'' \emph{International Journal of Intelligent Systems}, 2022.

\bibitem{zeng2022federated}
T.~Zeng, O.~Semiari, M.~Chen, W.~Saad, and M.~Bennis, ``Federated learning on the road autonomous controller design for connected and autonomous vehicles,'' \emph{IEEE Transactions on Wireless Communications}, vol.~21, no.~12, pp. 10\,407--10\,423, 2022.

\bibitem{yuan2023federated}
L.~Yuan, L.~Su, and Z.~Wang, ``Federated transfer-ordered-personalized learning for driver monitoring application,'' \emph{arXiv preprint arXiv:2301.04829}, 2023.

\bibitem{zhao2023fedsup}
C.~Zhao, Z.~Gao, Q.~Wang, K.~Xiao, Z.~Mo, and M.~J. Deen, ``Fedsup: A communication-efficient federated learning fatigue driving behaviors supervision approach,'' \emph{Future Generation Computer Systems}, vol. 138, pp. 52--60, 2023.

\bibitem{liu2019flownet3d}
X.~Liu, C.~R. Qi, and L.~J. Guibas, ``Flownet3d: Learning scene flow in 3d point clouds,'' in \emph{Proceedings of the IEEE/CVF Conference on Computer Vision and Pattern Recognition}, 2019, pp. 529--537.

\bibitem{fedavg}
B.~McMahan, E.~Moore, D.~Ramage, S.~Hampson, and B.~A. y~Arcas, ``Communication-efficient learning of deep networks from decentralized data,'' in \emph{Artificial intelligence and statistics}.\hskip 1em plus 0.5em minus 0.4em\relax PMLR, 2017, pp. 1273--1282.

\bibitem{fedavgm}
T.-M.~H. Hsu, H.~Qi, and M.~Brown, ``Measuring the effects of non-identical data distribution for federated visual classification,'' \emph{arXiv preprint arXiv:1909.06335}, 2019.

\bibitem{fedprox}
T.~Li, A.~K. Sahu, M.~Zaheer, M.~Sanjabi, A.~Talwalkar, and V.~Smith, ``Federated optimization in heterogeneous networks,'' \emph{Proceedings of Machine Learning and Systems}, vol.~2, pp. 429--450, 2020.

\bibitem{scaffold}
S.~P. Karimireddy, S.~Kale, M.~Mohri, S.~Reddi, S.~Stich, and A.~T. Suresh, ``Scaffold: Stochastic controlled averaging for federated learning,'' in \emph{International Conference on Machine Learning}.\hskip 1em plus 0.5em minus 0.4em\relax PMLR, 2020, pp. 5132--5143.

\bibitem{feddyn}
D.~A.~E. Acar, Y.~Zhao, R.~Matas, M.~Mattina, P.~Whatmough, and V.~Saligrama, ``Federated learning based on dynamic regularization,'' in \emph{International Conference on Learning Representations}, 2020.

\bibitem{fednova}
J.~Wang, Q.~Liu, H.~Liang, G.~Joshi, and H.~V. Poor, ``Tackling the objective inconsistency problem in heterogeneous federated optimization,'' \emph{Advances in neural information processing systems}, vol.~33, pp. 7611--7623, 2020.

\bibitem{fedopt}
\BIBentryALTinterwordspacing
S.~J. Reddi, Z.~Charles, M.~Zaheer, Z.~Garrett, K.~Rush, J.~Kone{\v{c}}n{\'y}, S.~Kumar, and H.~B. McMahan, ``Adaptive federated optimization,'' in \emph{International Conference on Learning Representations}, 2021. [Online]. Available: \url{https://openreview.net/forum?id=LkFG3lB13U5}
\BIBentrySTDinterwordspacing

\bibitem{ditto}
T.~Li, S.~Hu, A.~Beirami, and V.~Smith, ``Ditto: Fair and robust federated learning through personalization,'' in \emph{International Conference on Machine Learning}.\hskip 1em plus 0.5em minus 0.4em\relax PMLR, 2021, pp. 6357--6368.

\bibitem{fedper}
M.~G. Arivazhagan, V.~Aggarwal, A.~K. Singh, and S.~Choudhary, ``Federated learning with personalization layers,'' \emph{arXiv preprint arXiv:1912.00818}, 2019.

\bibitem{fedrep}
L.~Collins, H.~Hassani, A.~Mokhtari, and S.~Shakkottai, ``Exploiting shared representations for personalized federated learning,'' in \emph{International Conference on Machine Learning}.\hskip 1em plus 0.5em minus 0.4em\relax PMLR, 2021, pp. 2089--2099.

\bibitem{pfedme}
C.~T~Dinh, N.~Tran, and J.~Nguyen, ``Personalized federated learning with moreau envelopes,'' \emph{Advances in Neural Information Processing Systems}, vol.~33, pp. 21\,394--21\,405, 2020.

\bibitem{pfedgraph}
R.~Ye, Z.~Ni, F.~Wu, S.~Chen, and Y.~Wang, ``Personalized federated learning with inferred collaboration graphs,'' in \emph{International Conference on Machine Learning}.\hskip 1em plus 0.5em minus 0.4em\relax PMLR, 2023, pp. 39\,801--39\,817.

\bibitem{nuscenes}
H.~Caesar, V.~Bankiti, A.~H. Lang, S.~Vora, V.~E. Liong, Q.~Xu, A.~Krishnan, Y.~Pan, G.~Baldan, and O.~Beijbom, ``nuscenes: A multimodal dataset for autonomous driving,'' in \emph{Proceedings of the IEEE/CVF conference on computer vision and pattern recognition}, 2020, pp. 11\,621--11\,631.

\bibitem{waymoopen}
P.~Sun, H.~Kretzschmar, X.~Dotiwalla, A.~Chouard, V.~Patnaik, P.~Tsui, J.~Guo, Y.~Zhou, Y.~Chai, B.~Caine \emph{et~al.}, ``Scalability in perception for autonomous driving: Waymo open dataset,'' in \emph{Proceedings of the IEEE/CVF conference on computer vision and pattern recognition}, 2020, pp. 2446--2454.

\bibitem{BEVHeight}
L.~Yang, K.~Yu, T.~Tang, J.~Li, K.~Yuan, L.~Wang, X.~Zhang, and P.~Chen, ``Bevheight: A robust framework for vision-based roadside 3d object detection,'' in \emph{Proceedings of the IEEE/CVF Conference on Computer Vision and Pattern Recognition}, 2023, pp. 21\,611--21\,620.

\bibitem{hu2023aerial}
Y.~Hu, S.~Fang, W.~Xie, and S.~Chen, ``Aerial monocular 3d object detection,'' \emph{IEEE Robotics and Automation Letters}, vol.~8, no.~4, pp. 1959--1966, 2023.

\bibitem{zimmer2023infradet3d}
W.~Zimmer, J.~Birkner, M.~Brucker, H.~T. Nguyen, S.~Petrovski, B.~Wang, and A.~C. Knoll, ``Infradet3d: Multi-modal 3d object detection based on roadside infrastructure camera and lidar sensors,'' \emph{arXiv preprint arXiv:2305.00314}, 2023.

\bibitem{V2X-Seq}
H.~Yu, W.~Yang, H.~Ruan, Z.~Yang, Y.~Tang, X.~Gao, X.~Hao, Y.~Shi, Y.~Pan, N.~Sun \emph{et~al.}, ``V2x-seq: A large-scale sequential dataset for vehicle-infrastructure cooperative perception and forecasting,'' in \emph{Proceedings of the IEEE/CVF Conference on Computer Vision and Pattern Recognition}, 2023, pp. 5486--5495.

\bibitem{V2X-Sim}
Y.~Li, D.~Ma, Z.~An, Z.~Wang, Y.~Zhong, S.~Chen, and C.~Feng, ``V2x-sim: Multi-agent collaborative perception dataset and benchmark for autonomous driving,'' \emph{IEEE Robotics and Automation Letters}, vol.~7, no.~4, pp. 10\,914--10\,921, 2022.

\bibitem{arnold2020cooperative}
E.~Arnold, M.~Dianati, R.~de~Temple, and S.~Fallah, ``Cooperative perception for 3d object detection in driving scenarios using infrastructure sensors,'' \emph{IEEE Transactions on Intelligent Transportation Systems}, vol.~23, no.~3, pp. 1852--1864, 2020.

\bibitem{zhang2023robust}
R.~Zhang, D.~Meng, L.~Bassett, S.~Shen, Z.~Zou, and H.~X. Liu, ``Robust roadside perception for autonomous driving: an annotation-free strategy with synthesized data,'' \emph{arXiv preprint arXiv:2306.17302}, 2023.

\bibitem{azuma1997survey}
R.~T. Azuma, ``A survey of augmented reality,'' \emph{Presence: teleoperators \& virtual environments}, vol.~6, no.~4, pp. 355--385, 1997.

\bibitem{creswell2018generative}
A.~Creswell, T.~White, V.~Dumoulin, K.~Arulkumaran, B.~Sengupta, and A.~A. Bharath, ``Generative adversarial networks: An overview,'' \emph{IEEE signal processing magazine}, vol.~35, no.~1, pp. 53--65, 2018.

\bibitem{wu2023efficient}
A.~Wu, P.~He, X.~Li, K.~Chen, S.~Ranka, and A.~Rangarajan, ``An efficient semi-automated scheme for infrastructure lidar annotation,'' \emph{arXiv preprint arXiv:2301.10732}, 2023.

\bibitem{huguet2007variational}
F.~Huguet and F.~Devernay, ``A variational method for scene flow estimation from stereo sequences,'' in \emph{2007 IEEE 11th International Conference on Computer Vision}.\hskip 1em plus 0.5em minus 0.4em\relax IEEE, 2007, pp. 1--7.

\bibitem{carceroni2002multi}
R.~L. Carceroni and K.~N. Kutulakos, ``Multi-view scene capture by surfel sampling: From video streams to non-rigid 3d motion, shape and reflectance,'' \emph{International Journal of Computer Vision}, vol.~49, pp. 175--214, 2002.

\bibitem{pons2005modelling}
J.-P. Pons, R.~Keriven, and O.~Faugeras, ``Modelling dynamic scenes by registering multi-view image sequences,'' in \emph{2005 IEEE Computer Society Conference on Computer Vision and Pattern Recognition (CVPR'05)}, vol.~2.\hskip 1em plus 0.5em minus 0.4em\relax IEEE, 2005, pp. 822--827.

\bibitem{vogel20153d}
C.~Vogel, K.~Schindler, and S.~Roth, ``3d scene flow estimation with a piecewise rigid scene model,'' \emph{International Journal of Computer Vision}, vol. 115, pp. 1--28, 2015.

\bibitem{brickwedde2019mono}
F.~Brickwedde, S.~Abraham, and R.~Mester, ``Mono-sf: Multi-view geometry meets single-view depth for monocular scene flow estimation of dynamic traffic scenes,'' in \emph{Proceedings of the IEEE/CVF International Conference on Computer Vision}, 2019, pp. 2780--2790.

\bibitem{rishav2020deeplidarflow}
R.~Rishav, R.~Battrawy, R.~Schuster, O.~Wasenm{\"u}ller, and D.~Stricker, ``Deeplidarflow: A deep learning architecture for scene flow estimation using monocular camera and sparse lidar,'' in \emph{2020 IEEE/RSJ International Conference on Intelligent Robots and Systems (IROS)}.\hskip 1em plus 0.5em minus 0.4em\relax IEEE, 2020, pp. 10\,460--10\,467.

\bibitem{shao2018motion}
L.~Shao, P.~Shah, V.~Dwaracherla, and J.~Bohg, ``Motion-based object segmentation based on dense rgb-d scene flow,'' \emph{IEEE Robotics and Automation Letters}, vol.~3, no.~4, pp. 3797--3804, 2018.

\bibitem{teed2021raft}
Z.~Teed and J.~Deng, ``Raft-3d: Scene flow using rigid-motion embeddings,'' in \emph{Proceedings of the IEEE/CVF Conference on Computer Vision and Pattern Recognition}, 2021, pp. 8375--8384.

\bibitem{pointnet}
C.~R. Qi, H.~Su, K.~Mo, and L.~J. Guibas, ``Pointnet: Deep learning on point sets for 3d classification and segmentation,'' in \emph{Proceedings of the IEEE conference on computer vision and pattern recognition}, 2017, pp. 652--660.

\bibitem{pointnet++}
C.~R. Qi, L.~Yi, H.~Su, and L.~J. Guibas, ``Pointnet++: Deep hierarchical feature learning on point sets in a metric space,'' \emph{Advances in neural information processing systems}, vol.~30, 2017.

\bibitem{lang2019pointpillars}
A.~H. Lang, S.~Vora, H.~Caesar, L.~Zhou, J.~Yang, and O.~Beijbom, ``Pointpillars: Fast encoders for object detection from point clouds,'' in \emph{Proceedings of the IEEE/CVF conference on computer vision and pattern recognition}, 2019, pp. 12\,697--12\,705.

\bibitem{puy2020flot}
G.~Puy, A.~Boulch, and R.~Marlet, ``Flot: Scene flow on point clouds guided by optimal transport,'' in \emph{Computer Vision--ECCV 2020: 16th European Conference, Glasgow, UK, August 23--28, 2020, Proceedings, Part XXVIII}.\hskip 1em plus 0.5em minus 0.4em\relax Springer, 2020, pp. 527--544.

\bibitem{gu2019hplflownet}
X.~Gu, Y.~Wang, C.~Wu, Y.~J. Lee, and P.~Wang, ``Hplflownet: Hierarchical permutohedral lattice flownet for scene flow estimation on large-scale point clouds,'' in \emph{Proceedings of the IEEE/CVF conference on computer vision and pattern recognition}, 2019, pp. 3254--3263.

\bibitem{fastflow3d}
P.~Jund, C.~Sweeney, N.~Abdo, Z.~Chen, and J.~Shlens, ``Scalable scene flow from point clouds in the real world,'' \emph{IEEE Robotics and Automation Letters}, vol.~7, no.~2, pp. 1589--1596, 2021.

\bibitem{cheng2022bi}
W.~Cheng and J.~H. Ko, ``Bi-pointflownet: Bidirectional learning for point cloud based scene flow estimation,'' in \emph{Computer Vision--ECCV 2022: 17th European Conference, Tel Aviv, Israel, October 23--27, 2022, Proceedings, Part XXVIII}.\hskip 1em plus 0.5em minus 0.4em\relax Springer, 2022, pp. 108--124.

\bibitem{li2021hcrf}
R.~Li, G.~Lin, T.~He, F.~Liu, and C.~Shen, ``Hcrf-flow: Scene flow from point clouds with continuous high-order crfs and position-aware flow embedding,'' in \emph{Proceedings of the IEEE/CVF Conference on Computer Vision and Pattern Recognition}, 2021, pp. 364--373.

\bibitem{li2022rigidflow}
R.~Li, C.~Zhang, G.~Lin, Z.~Wang, and C.~Shen, ``Rigidflow: Self-supervised scene flow learning on point clouds by local rigidity prior,'' in \emph{Proceedings of the IEEE/CVF Conference on Computer Vision and Pattern Recognition}, 2022, pp. 16\,959--16\,968.

\bibitem{tishchenko2020self}
I.~Tishchenko, S.~Lombardi, M.~R. Oswald, and M.~Pollefeys, ``Self-supervised learning of non-rigid residual flow and ego-motion,'' in \emph{2020 international conference on 3D vision (3DV)}.\hskip 1em plus 0.5em minus 0.4em\relax IEEE, 2020, pp. 150--159.

\bibitem{sorkine2005laplacian}
O.~Sorkine, ``Laplacian mesh processing,'' \emph{Eurographics (State of the Art Reports)}, vol.~4, no.~4, 2005.

\bibitem{dong2022exploiting}
G.~Dong, Y.~Zhang, H.~Li, X.~Sun, and Z.~Xiong, ``Exploiting rigidity constraints for lidar scene flow estimation,'' in \emph{Proceedings of the IEEE/CVF Conference on Computer Vision and Pattern Recognition}, 2022, pp. 12\,776--12\,785.

\bibitem{fed_empirical}
Y.~Zhao, M.~Li, L.~Lai, N.~Suda, D.~Civin, and V.~Chandra, ``Federated learning with non-iid data,'' \emph{arXiv preprint arXiv:1806.00582}, 2018.

\bibitem{li2019convergence}
X.~Li, K.~Huang, W.~Yang, S.~Wang, and Z.~Zhang, ``On the convergence of fedavg on non-iid data,'' in \emph{International Conference on Learning Representations}, 2019.

\bibitem{wang2021cooperative}
J.~Wang and G.~Joshi, ``Cooperative sgd: A unified framework for the design and analysis of local-update sgd algorithms,'' \emph{The Journal of Machine Learning Research}, vol.~22, no.~1, pp. 9709--9758, 2021.

\bibitem{flair}
C.~Song, F.~Granqvist, and K.~Talwar, ``Flair: Federated learning annotated image repository,'' in \emph{Thirty-sixth Conference on Neural Information Processing Systems Datasets and Benchmarks Track}, 2022.

\bibitem{flamby}
J.~O. du~Terrail, S.-S. Ayed, E.~Cyffers, F.~Grimberg, C.~He, R.~Loeb, P.~Mangold, T.~Marchand, O.~Marfoq, E.~Mushtaq \emph{et~al.}, ``Flamby: Datasets and benchmarks for cross-silo federated learning in realistic healthcare settings,'' in \emph{Thirty-sixth Conference on Neural Information Processing Systems Datasets and Benchmarks Track}, 2022.

\bibitem{leaf}
S.~Caldas, S.~M.~K. Duddu, P.~Wu, T.~Li, J.~Kone{\v{c}}n{\`y}, H.~B. McMahan, V.~Smith, and A.~Talwalkar, ``Leaf: A benchmark for federated settings,'' \emph{arXiv preprint arXiv:1812.01097}, 2018.

\bibitem{pflbench}
D.~Chen, D.~Gao, W.~Kuang, Y.~Li, and B.~Ding, ``pfl-bench: A comprehensive benchmark for personalized federated learning,'' \emph{Advances in Neural Information Processing Systems}, vol.~35, pp. 9344--9360, 2022.

\bibitem{moon}
Q.~Li, B.~He, and D.~Song, ``Model-contrastive federated learning,'' in \emph{Proceedings of the IEEE/CVF Conference on Computer Vision and Pattern Recognition}, 2021, pp. 10\,713--10\,722.

\bibitem{fedfm}
R.~Ye, Z.~Ni, C.~Xu, J.~Wang, S.~Chen, and Y.~C. Eldar, ``Fedfm: Anchor-based feature matching for data heterogeneity in federated learning,'' \emph{IEEE Transactions on Signal Processing}, 2023.

\bibitem{vrlsgd}
X.~Liang, S.~Shen, J.~Liu, Z.~Pan, E.~Chen, and Y.~Cheng, ``Variance reduced local sgd with lower communication complexity,'' \emph{arXiv preprint arXiv:1912.12844}, 2019.

\bibitem{feddf}
T.~Lin, L.~Kong, S.~U. Stich, and M.~Jaggi, ``Ensemble distillation for robust model fusion in federated learning,'' \emph{Advances in Neural Information Processing Systems}, vol.~33, pp. 2351--2363, 2020.

\bibitem{fedgen}
Z.~Zhu, J.~Hong, and J.~Zhou, ``Data-free knowledge distillation for heterogeneous federated learning,'' in \emph{International Conference on Machine Learning}.\hskip 1em plus 0.5em minus 0.4em\relax PMLR, 2021, pp. 12\,878--12\,889.

\bibitem{feddisco}
R.~Ye, M.~Xu, J.~Wang, C.~Xu, S.~Chen, and Y.~Wang, ``Feddisco: Federated learning with discrepancy-aware collaboration,'' \emph{arXiv preprint arXiv:2305.19229}, 2023.

\bibitem{per_ml_1}
Y.~Jiang, J.~Kone{\v{c}}n{\`y}, K.~Rush, and S.~Kannan, ``Improving federated learning personalization via model agnostic meta learning,'' \emph{arXiv preprint arXiv:1909.12488}, 2019.

\bibitem{per_ml_2}
A.~Fallah, A.~Mokhtari, and A.~Ozdaglar, ``Personalized federated learning with theoretical guarantees: A model-agnostic meta-learning approach,'' \emph{Advances in Neural Information Processing Systems}, vol.~33, pp. 3557--3568, 2020.

\bibitem{cfl}
F.~Sattler, K.-R. M{\"u}ller, and W.~Samek, ``Clustered federated learning: Model-agnostic distributed multitask optimization under privacy constraints,'' \emph{IEEE transactions on neural networks and learning systems}, vol.~32, no.~8, pp. 3710--3722, 2020.

\bibitem{fedma}
\BIBentryALTinterwordspacing
H.~Wang, M.~Yurochkin, Y.~Sun, D.~Papailiopoulos, and Y.~Khazaeni, ``Federated learning with matched averaging,'' in \emph{International Conference on Learning Representations}, 2020. [Online]. Available: \url{https://openreview.net/forum?id=BkluqlSFDS}
\BIBentrySTDinterwordspacing

\bibitem{bayesian}
M.~Yurochkin, M.~Agarwal, S.~Ghosh, K.~Greenewald, N.~Hoang, and Y.~Khazaeni, ``Bayesian nonparametric federated learning of neural networks,'' in \emph{International Conference on Machine Learning}.\hskip 1em plus 0.5em minus 0.4em\relax PMLR, 2019, pp. 7252--7261.

\bibitem{mnist}
Y.~LeCun, C.~Cortes, and C.~Burges, ``Mnist handwritten digit database,'' \emph{ATT Labs [Online]. Available: http://yann.lecun.com/exdb/mnist}, vol.~2, 2010.

\bibitem{xiao2017fashion}
H.~Xiao, K.~Rasul, and R.~Vollgraf, ``Fashion-mnist: a novel image dataset for benchmarking machine learning algorithms,'' \emph{arXiv preprint arXiv:1708.07747}, 2017.

\bibitem{cifar10}
A.~Krizhevsky, G.~Hinton \emph{et~al.}, ``Learning multiple layers of features from tiny images,'' 2009.

\bibitem{darlow2018cinic}
L.~N. Darlow, E.~J. Crowley, A.~Antoniou, and A.~J. Storkey, ``Cinic-10 is not imagenet or cifar-10,'' \emph{arXiv preprint arXiv:1810.03505}, 2018.

\bibitem{imagenet}
J.~Deng, W.~Dong, R.~Socher, L.-J. Li, K.~Li, and L.~Fei-Fei, ``Imagenet: A large-scale hierarchical image database,'' in \emph{2009 IEEE conference on computer vision and pattern recognition}.\hskip 1em plus 0.5em minus 0.4em\relax IEEE, 2009, pp. 248--255.

\bibitem{fed_da}
X.~Peng, Z.~Huang, Y.~Zhu, and K.~Saenko, ``Federated adversarial domain adaptation,'' in \emph{International Conference on Learning Representations}, 2019.

\bibitem{fedbn}
X.~Li, M.~JIANG, X.~Zhang, M.~Kamp, and Q.~Dou, ``Fedbn: Federated learning on non-iid features via local batch normalization,'' in \emph{International Conference on Learning Representations}, 2020.

\bibitem{fedsr}
A.~T. Nguyen, P.~Torr, and S.~N. Lim, ``Fedsr: A simple and effective domain generalization method for federated learning,'' \emph{Advances in Neural Information Processing Systems}, vol.~35, pp. 38\,831--38\,843, 2022.

\bibitem{svhn}
Y.~Netzer, T.~Wang, A.~Coates, A.~Bissacco, B.~Wu, and A.~Y. Ng, ``Reading digits in natural images with unsupervised feature learning,'' 2011.

\bibitem{office-home}
H.~Venkateswara, J.~Eusebio, S.~Chakraborty, and S.~Panchanathan, ``Deep hashing network for unsupervised domain adaptation,'' in \emph{Proceedings of the IEEE Conference on Computer Vision and Pattern Recognition}, 2017, pp. 5018--5027.

\bibitem{domainnet}
X.~Peng, Q.~Bai, X.~Xia, Z.~Huang, K.~Saenko, and B.~Wang, ``Moment matching for multi-source domain adaptation,'' in \emph{Proceedings of the IEEE International Conference on Computer Vision}, 2019, pp. 1406--1415.

\bibitem{what}
H.~Yuan, W.~R. Morningstar, L.~Ning, and K.~Singhal, ``What do we mean by generalization in federated learning?'' in \emph{International Conference on Learning Representations}, 2021.

\bibitem{sentiment140}
A.~Go, R.~Bhayani, and L.~Huang, ``Twitter sentiment classification using distant supervision,'' \emph{CS224N project report, Stanford}, vol.~1, no.~12, p. 2009, 2009.

\bibitem{inaturalist}
G.~Van~Horn, O.~Mac~Aodha, Y.~Song, Y.~Cui, C.~Sun, A.~Shepard, H.~Adam, P.~Perona, and S.~Belongie, ``The inaturalist species classification and detection dataset,'' in \emph{Proceedings of the IEEE conference on computer vision and pattern recognition}, 2018, pp. 8769--8778.

\bibitem{landmark}
T.~Weyand, A.~Araujo, B.~Cao, and J.~Sim, ``Google landmarks dataset v2-a large-scale benchmark for instance-level recognition and retrieval,'' in \emph{Proceedings of the IEEE/CVF conference on computer vision and pattern recognition}, 2020, pp. 2575--2584.

\bibitem{fedml}
C.~He, S.~Li, J.~So, X.~Zeng, M.~Zhang, H.~Wang, X.~Wang, P.~Vepakomma, A.~Singh, H.~Qiu \emph{et~al.}, ``Fedml: A research library and benchmark for federated machine learning,'' \emph{arXiv preprint arXiv:2007.13518}, 2020.

\bibitem{fedscale}
F.~Lai, Y.~Dai, S.~Singapuram, J.~Liu, X.~Zhu, H.~Madhyastha, and M.~Chowdhury, ``Fedscale: Benchmarking model and system performance of federated learning at scale,'' in \emph{International Conference on Machine Learning}.\hskip 1em plus 0.5em minus 0.4em\relax PMLR, 2022, pp. 11\,814--11\,827.

\bibitem{fednlp}
B.~Y. Lin, C.~He, Z.~Zeng, H.~Wang, Y.~Huang, M.~Soltanolkotabi, X.~Ren, and S.~Avestimehr, ``Fednlp: A research platform for federated learning in natural language processing,'' \emph{arXiv preprint arXiv:2104.08815}, 2021.

\bibitem{federatedscope}
Z.~Wang, W.~Kuang, Y.~Xie, L.~Yao, Y.~Li, B.~Ding, and J.~Zhou, ``Federatedscope-gnn: Towards a unified, comprehensive and efficient package for federated graph learning,'' in \emph{Proceedings of the 28th ACM SIGKDD Conference on Knowledge Discovery and Data Mining}, 2022, pp. 4110--4120.

\bibitem{song2022ogc}
Z.~Song and B.~Yang, ``Ogc: Unsupervised 3d object segmentation from rigid dynamics of point clouds,'' \emph{Advances in Neural Information Processing Systems}, vol.~35, pp. 30\,798--30\,812, 2022.

\bibitem{najibi2022motion}
M.~Najibi, J.~Ji, Y.~Zhou, C.~R. Qi, X.~Yan, S.~Ettinger, and D.~Anguelov, ``Motion inspired unsupervised perception and prediction in autonomous driving,'' in \emph{European Conference on Computer Vision}.\hskip 1em plus 0.5em minus 0.4em\relax Springer, 2022, pp. 424--443.

\bibitem{teed2020raft}
Z.~Teed and J.~Deng, ``Raft: Recurrent all-pairs field transforms for optical flow,'' in \emph{Computer Vision--ECCV 2020: 16th European Conference, Glasgow, UK, August 23--28, 2020, Proceedings, Part II 16}.\hskip 1em plus 0.5em minus 0.4em\relax Springer, 2020, pp. 402--419.

\bibitem{paszke2019pytorch}
A.~Paszke, S.~Gross, F.~Massa, A.~Lerer, J.~Bradbury, G.~Chanan, T.~Killeen, Z.~Lin, N.~Gimelshein, L.~Antiga \emph{et~al.}, ``Pytorch: An imperative style, high-performance deep learning library,'' \emph{Advances in neural information processing systems}, vol.~32, 2019.

\bibitem{lifair}
T.~Li, M.~Sanjabi, A.~Beirami, and V.~Smith, ``Fair resource allocation in federated learning,'' in \emph{International Conference on Learning Representations}, 2019.

\bibitem{fedala}
J.~Zhang, Y.~Hua, H.~Wang, T.~Song, Z.~Xue, R.~Ma, and H.~Guan, ``Fedala: Adaptive local aggregation for personalized federated learning,'' in \emph{Proceedings of the AAAI Conference on Artificial Intelligence}, vol.~37, no.~9, 2023, pp. 11\,237--11\,244.

\bibitem{fedrod}
H.-Y. Chen and W.-L. Chao, ``On bridging generic and personalized federated learning for image classification,'' in \emph{International Conference on Learning Representations}, 2021.

\bibitem{clip}
A.~Radford, J.~W. Kim, C.~Hallacy, A.~Ramesh, G.~Goh, S.~Agarwal, G.~Sastry, A.~Askell, P.~Mishkin, J.~Clark \emph{et~al.}, ``Learning transferable visual models from natural language supervision,'' in \emph{International conference on machine learning}.\hskip 1em plus 0.5em minus 0.4em\relax PMLR, 2021, pp. 8748--8763.

\bibitem{gpt4}
OpenAI, ``Gpt-4 technical report,'' \emph{ArXiv}, vol. abs/2303.08774, 2023.

\bibitem{li2022simipu}
Z.~Li, Z.~Chen, A.~Li, L.~Fang, Q.~Jiang, X.~Liu, J.~Jiang, B.~Zhou, and H.~Zhao, ``Simipu: Simple 2d image and 3d point cloud unsupervised pre-training for spatial-aware visual representations,'' in \emph{Proceedings of the AAAI Conference on Artificial Intelligence}, vol.~36, no.~2, 2022, pp. 1500--1508.

\bibitem{min2022voxel}
C.~Min, D.~Zhao, L.~Xiao, Y.~Nie, and B.~Dai, ``Voxel-mae: Masked autoencoders for pre-training large-scale point clouds,'' \emph{arXiv preprint arXiv:2206.09900}, 2022.

\bibitem{zhang2023towards}
L.~Zhang, A.~J. Yang, Y.~Xiong, S.~Casas, B.~Yang, M.~Ren, and R.~Urtasun, ``Towards unsupervised object detection from lidar point clouds,'' in \emph{Proceedings of the IEEE/CVF Conference on Computer Vision and Pattern Recognition}, 2023, pp. 9317--9328.

\end{thebibliography}

\begin{IEEEbiography}
[{\includegraphics[width=1in,height=1.25in, clip,keepaspectratio]{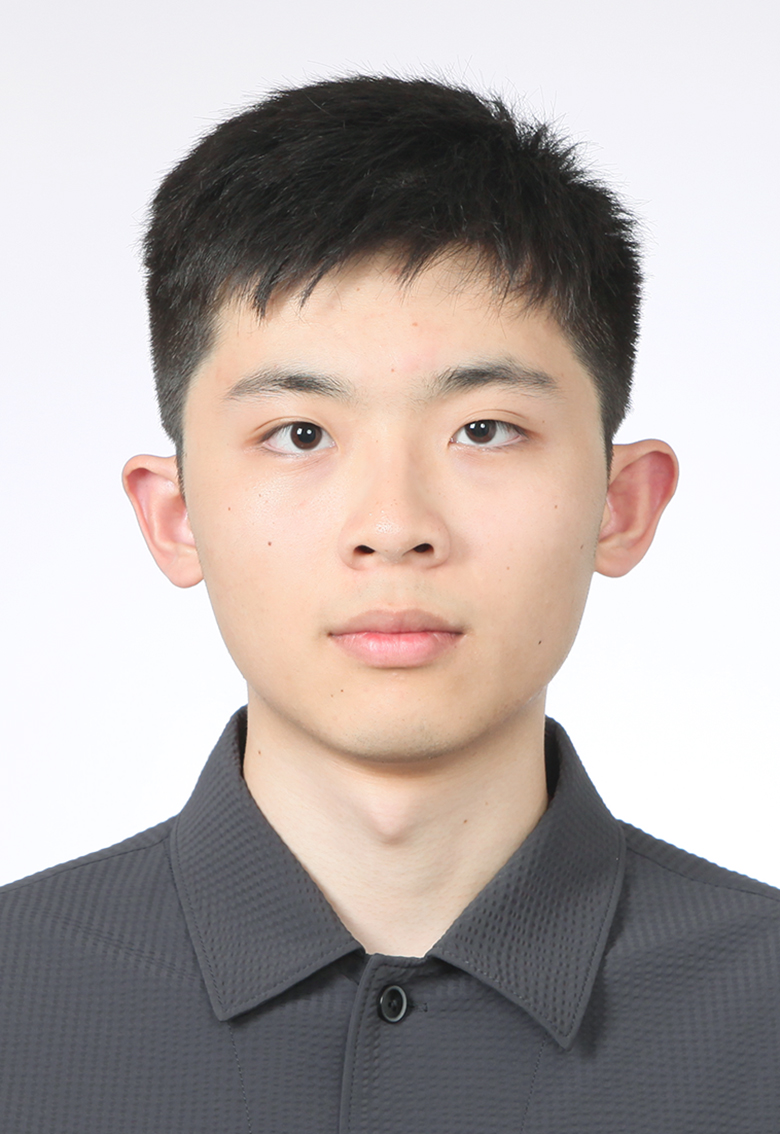}}]{Shaoheng Fang} received the B.E. degree in Computer Science from Shanghai Jiao Tong University, Shanghai, China, in 2022. 
He is currently pursuing the master’s degree with
the Department of Computer Science, University of Texas at Austin. 
He was a Research Intern at the Cooperative Medianet Innovation Center in Shanghai Jiao Tong University from 2021 to 2023. 
His research interests include autonomous driving, computer vision, and generative AI. 
\end{IEEEbiography}

\begin{IEEEbiography}
[{\includegraphics[width=1in,height=1.25in, clip,keepaspectratio]{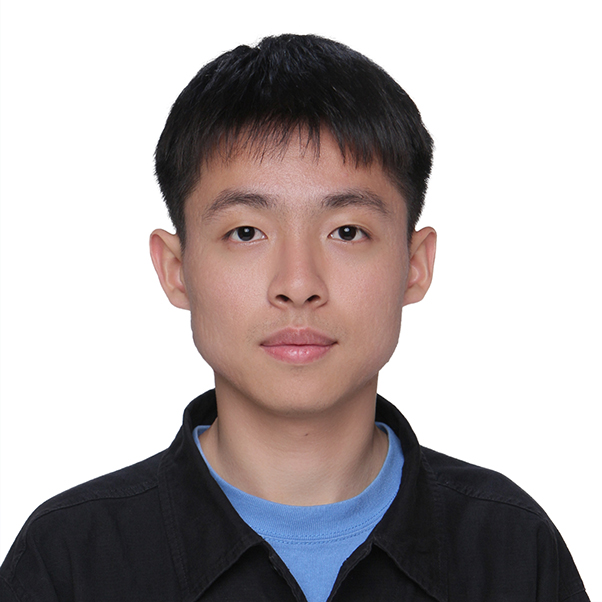}}]{Rui Ye} received the B.E. degree in Information Engineering from Shanghai Jiao Tong University, Shanghai, China, in 2022. He is currently working toward a Ph.D. degree at the Cooperative Medianet Innovation Center in Shanghai Jiao Tong University since 2022. He was a Research Intern with Microsoft Research Asia in 2022 \& 2023. His research interests include federated learning, responsible AI (e.g., trustworthy large language models), and multi-agent collaboration. 
\end{IEEEbiography}

\begin{IEEEbiography}
[{\includegraphics[width=1in,height=1.25in, clip,keepaspectratio]{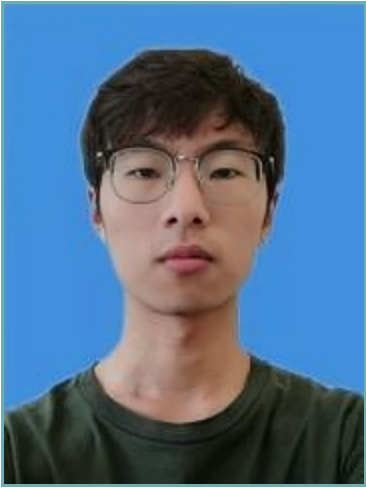}}]{Wenhao Wang} received the B.S. degree in Cyberspace Security from Wuhan University, Wuhan, China, in 2023. He is currently working toward the Ph.D. degree in the department of Computer Science with Zhejiang University, Hangzhou, China. He is now also an intern with Shanghai AI Laboratory. His research interests include federated learning, large language models and computer vision.
\end{IEEEbiography}

\begin{IEEEbiography}
[{\includegraphics[width=1in,height=1.25in, clip,keepaspectratio]{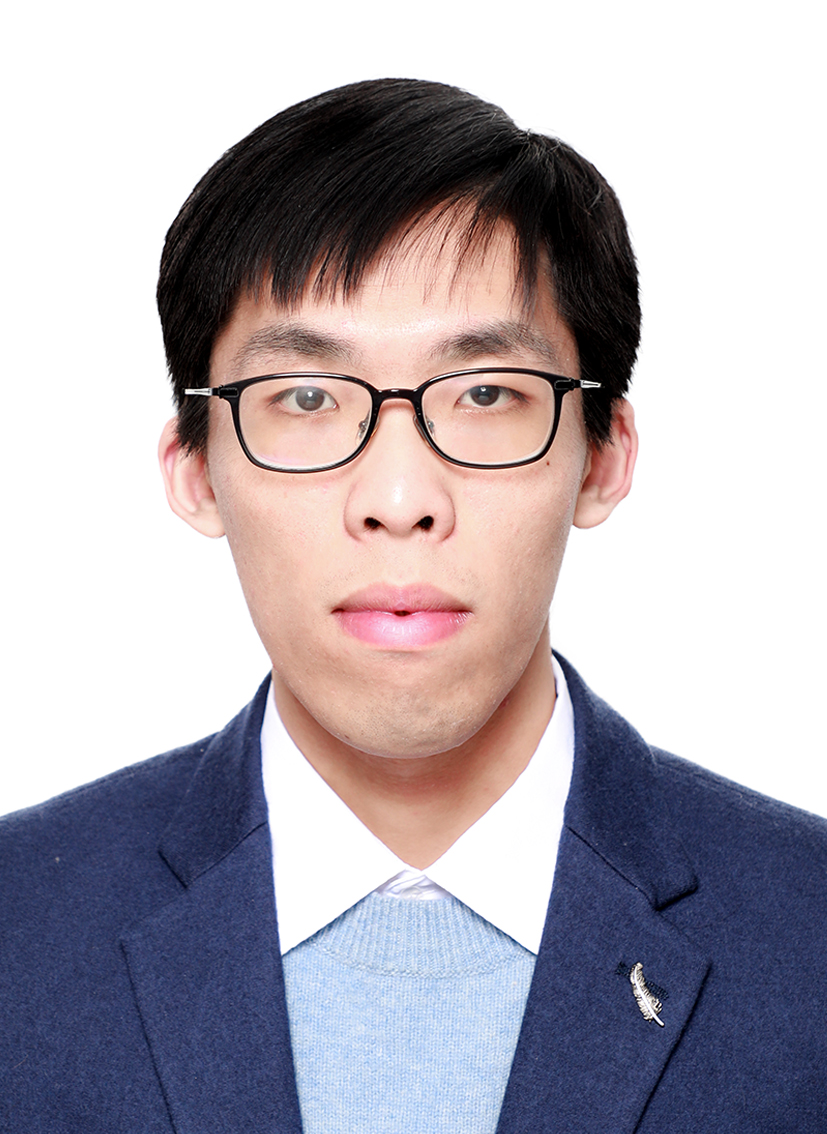}}]{Zuhong Liu} received the bachelor's degree in Information Engineering from SPEIT, Shanghai Jiao Tong University, Shanghai, China, in 2023. He is currently pursuing the master’s degree (a double degree program) with Shanghai Jiao Tong
University and Ecole Polytechnique, Palaiseau, France. His research interests mainly include autonomous driving and multi-agent collaboration.
\end{IEEEbiography}

\begin{IEEEbiography}
[{\includegraphics[width=1in,height=1.25in, clip,keepaspectratio]{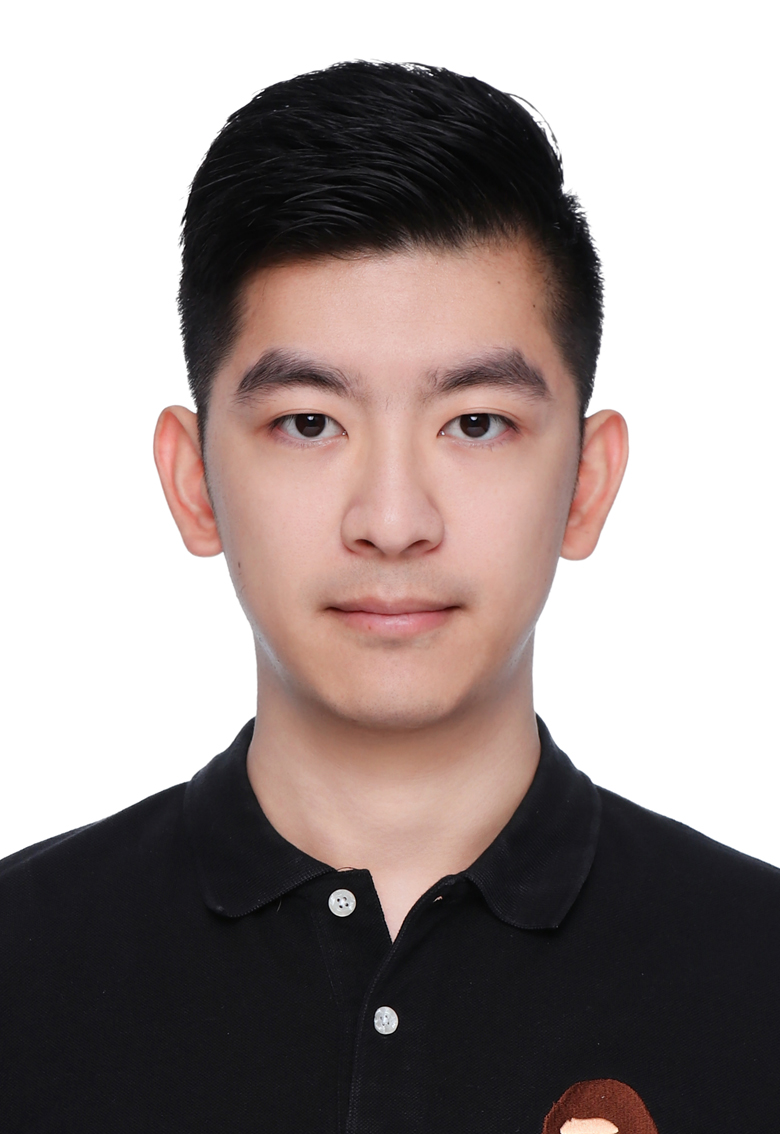}}]{Yuxiao Wang} received the B.A. degree in Computer Science from Boston University, MA, United State. He previously worked in Amazon, Seattle as Software Engineer and in NoiseMaker as Engineering Technical Lead. His research interests mainly include federated learning.
\end{IEEEbiography}

\begin{IEEEbiography}[{\includegraphics[width=1in,height=1.25in,clip,keepaspectratio]{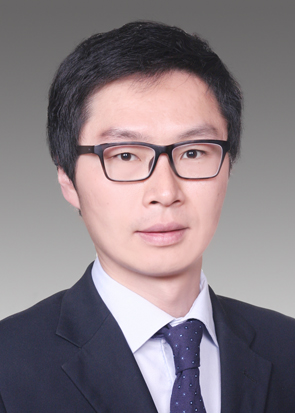}}]{Yafei Wang} (Member, IEEE) received the B.S. degree in internal combustion engine from Jilin University, Changchun, China, in 2005, the M.S. degree in vehicle engineering from Shanghai Jiao Tong University, Shanghai, China, in 2008, and the Ph.D. degree in electrical engineering from The University of Tokyo, Tokyo, Japan, in 2013.
From 2008 to 2010, he was with automotive industry for nearly two years. From 2013 to 2016, he was a Postdoctoral Researcher with The University of Tokyo. He is currently a Associate Professor with the School of Mechanical Engineering, Shanghai Jiao Tong University. His research interests include state estimation and control for connected and automated
vehicles
\end{IEEEbiography}

\begin{IEEEbiography}
[{\includegraphics[width=1in,height=1.25in, clip,keepaspectratio]{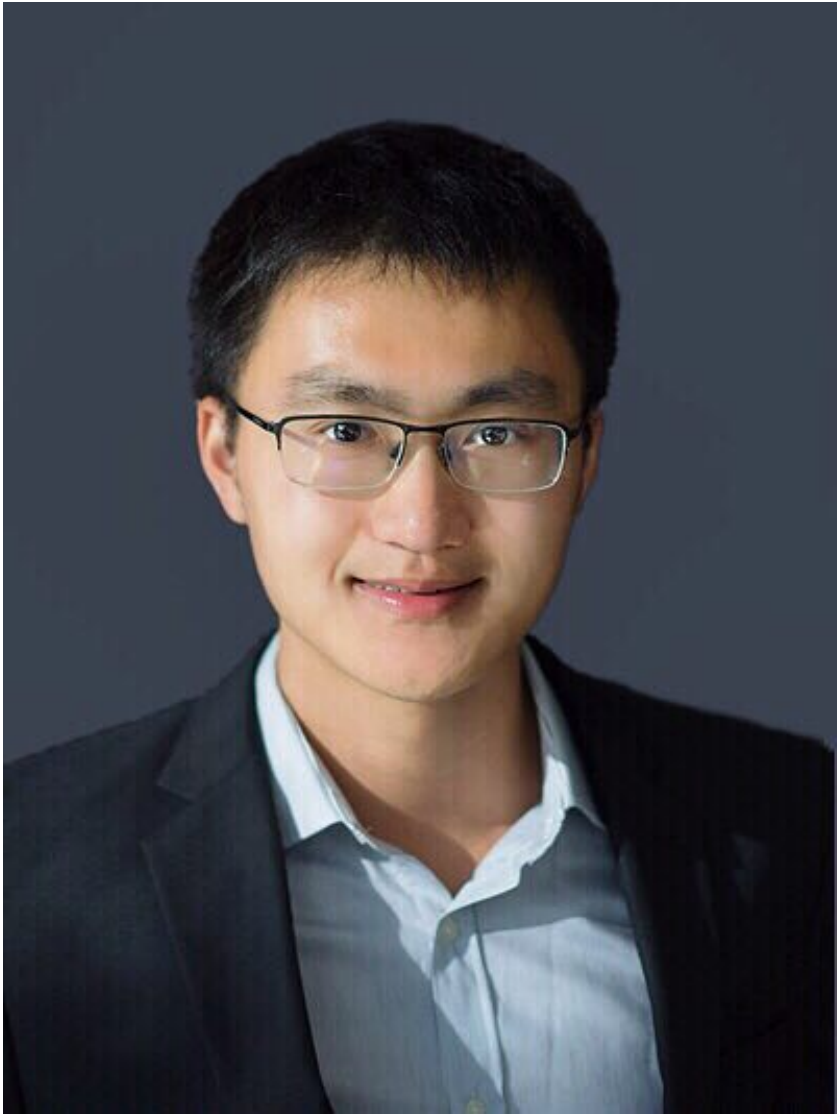}}]{Siheng Chen} (Senior Member, IEEE) is a tenure-track associate professor of Shanghai Jiao Tong University. Before joining Shanghai Jiao Tong University, he was a research scientist at Mitsubishi Electric Research Laboratories (MERL), and an autonomy engineer at Uber Advanced Technologies Group (ATG), working on the perception and prediction systems of self-driving cars. Before joining the industry, Dr. Chen was a postdoctoral research associate at Carnegie Mellon University. Dr. Chen received his doctorate in Electrical and Computer Engineering from Carnegie Mellon University in 2016, where he also received two master's degrees in Electrical and Computer Engineering (College of Engineering) and Machine Learning (School of Computer Science), respectively. He received his bachelor’s degree in Electronics Engineering in 2011 from the Beijing Institute of Technology, China. Dr. Chen's work on the sampling theory of graph data received the 2018 IEEE Signal Processing Society Young Author Best Paper Award. His co-authored paper on structural health monitoring received ASME SHM/NDE 2020 Best Journal Paper Runner-Up Award and another paper on 3D point cloud processing received the Best Student Paper Award at the 2018 IEEE Global Conference on Signal and Information Processing. Dr. Chen contributed to the project of scene-aware interaction, winning MERL President's Award. His research interests include graph neural networks, autonomous driving, and collective intelligence.
\end{IEEEbiography}

\begin{IEEEbiography}[{\includegraphics[width=1in,height=1.25in,clip,keepaspectratio]{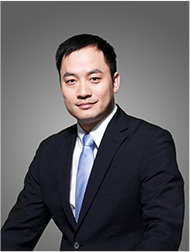}}]{Yanfeng Wang} received the B.E. degree in information engineering from the University of PLA, Beijing, China, and the M.S. and Ph.D. degrees in business management from the Antai College of Economics and Management, Shanghai Jiao Tong University, Shanghai, China. He is currently the Vice Director of the Cooperative Medianet Innovation Center and also the Vice Dean of the School of Electrical and Information Engineering, Shanghai Jiao Tong University. His research interests mainly include media big data and emerging commercial applications of information technology.
\end{IEEEbiography}

\end{document}


\appendix

\section{Datatset}

\begin{figure}[h]
    \centering
    \includegraphics[width=0.8\columnwidth]{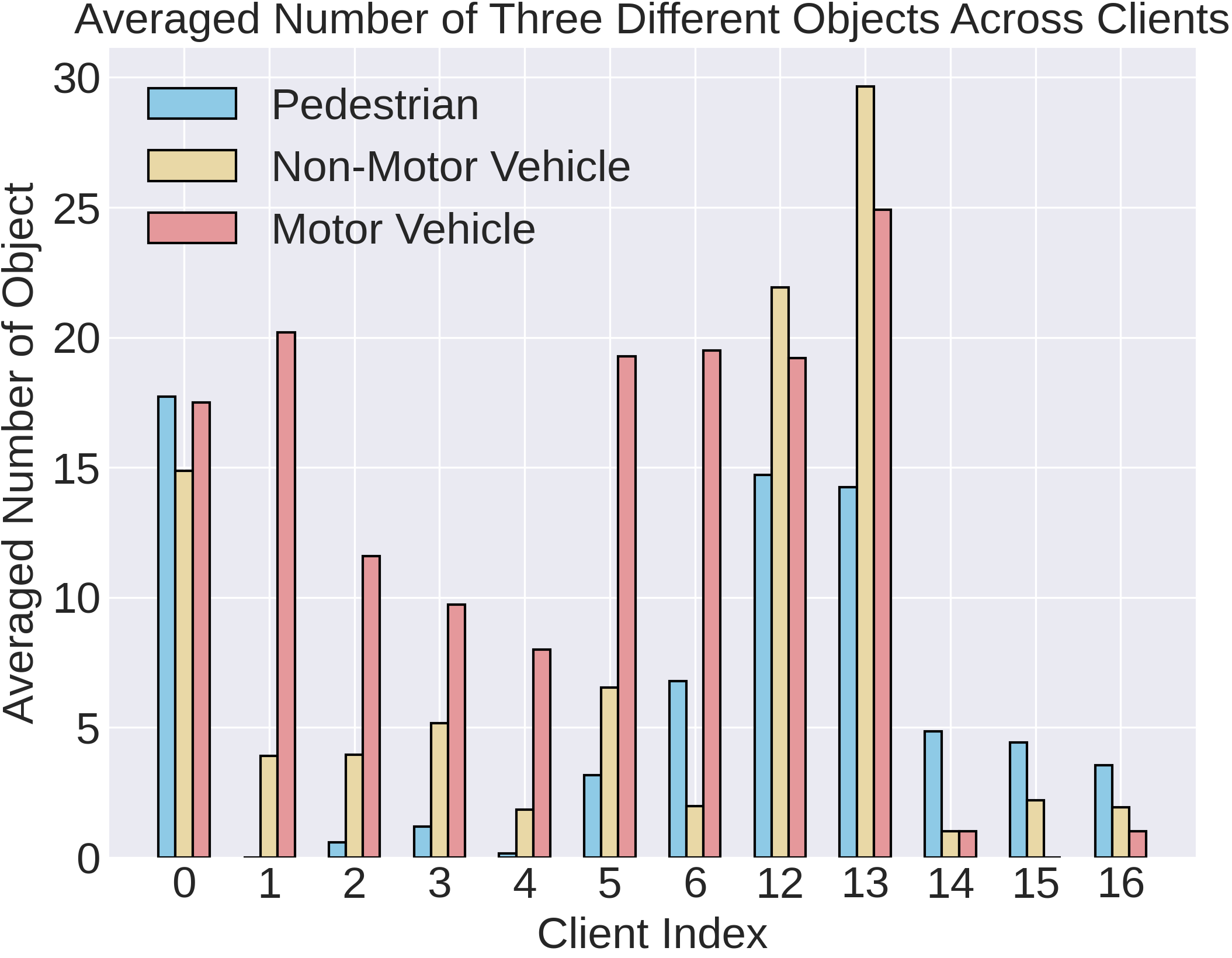}
    \caption{Client-level statistics of the number of three object categories. This indicates that the frequency of objects for each client can differ significantly.}
    \label{fig:data_label_client}
\end{figure}